\documentclass[10pt,twocolumn,letterpaper]{article}
\usepackage{cvpr}              

\usepackage{graphicx}
\usepackage{amsmath}
\usepackage{amssymb}
\usepackage{balance}
\usepackage{booktabs}
\usepackage{threeparttable}
\usepackage{multirow}
\usepackage{makecell}
\usepackage{setspace}

\usepackage[pagebackref,breaklinks,colorlinks]{hyperref}
\hypersetup{
    colorlinks=true,
    linkcolor=red,
    citecolor=cyan,
    filecolor=magenta,      
    urlcolor=red,
    }
\usepackage[export]{adjustbox} 
\usepackage{array,multirow}
\usepackage[accsupp]{axessibility}
\usepackage[capitalize]{cleveref}
\crefname{section}{Section}{Sections.}
\Crefname{section}{Section}{Sections}
\Crefname{table}{Table}{Tables}
\crefname{table}{Table}{Table}
\Crefname{figure}{Figure}{Figures}
\crefname{figure}{Figure}{Figure}

\def\cvprPaperID{12597} 
\def\confName{CVPR}
\def\confYear{2023}

\begin{document}

\title{Disentangling Writer and Character Styles for Handwriting Generation}

\author{Gang Dai$^{1}$\thanks{Authors contributed equally.}, Yifan Zhang$^{2*}$, Qingfeng Wang$^{1}$, Qing Du$^{1}$, 
Zhuliang Yu$^{1}$, \\ Zhuoman Liu$^{3}$, Shuangping Huang$^{1,4}$\thanks{Corresponding author} \\
$^{1}$South China University of Technology,
$^{2}$National University of Singapore, \\
$^{3}$The Hong Kong Polytechnic University,
$^{4}$Pazhou Laboratory \\
{\tt\small eedaigang@mail.scut.edu.cn}, {\tt\small yifan.zhang@u.nus.edu}, {\tt\small eehsp@scut.edu.cn}
}
\maketitle

\begin{abstract}
   Training machines to synthesize diverse handwritings is an intriguing task. Recently, RNN-based methods have been proposed to generate stylized online Chinese characters. However, these methods mainly focus on capturing a person’s overall writing style, neglecting subtle style inconsistencies between characters written by the same person. For example, while a person's handwriting typically exhibits general uniformity (\eg, glyph slant and aspect ratios), there are still small style variations in finer details (\eg, stroke length and curvature) of characters. In light of this, we propose to disentangle the style representations at both writer and character levels from individual handwritings to synthesize realistic stylized online handwritten characters. Specifically, we present the style-disentangled Transformer (SDT), which employs two complementary contrastive objectives to extract the style commonalities of reference samples and capture the detailed style patterns of each sample, respectively. Extensive experiments on various language scripts demonstrate the effectiveness of SDT. Notably, our empirical findings reveal that the two learned style representations provide information at different frequency magnitudes, underscoring the importance of separate style extraction. Our source code is public at: \url{https://github.com/dailenson/SDT}.
\end{abstract}
\vspace{-0.1in}

\section{Introduction}
\label{sec:intro}

\begin{figure}[t]
\centering
\includegraphics[width=0.83\linewidth]{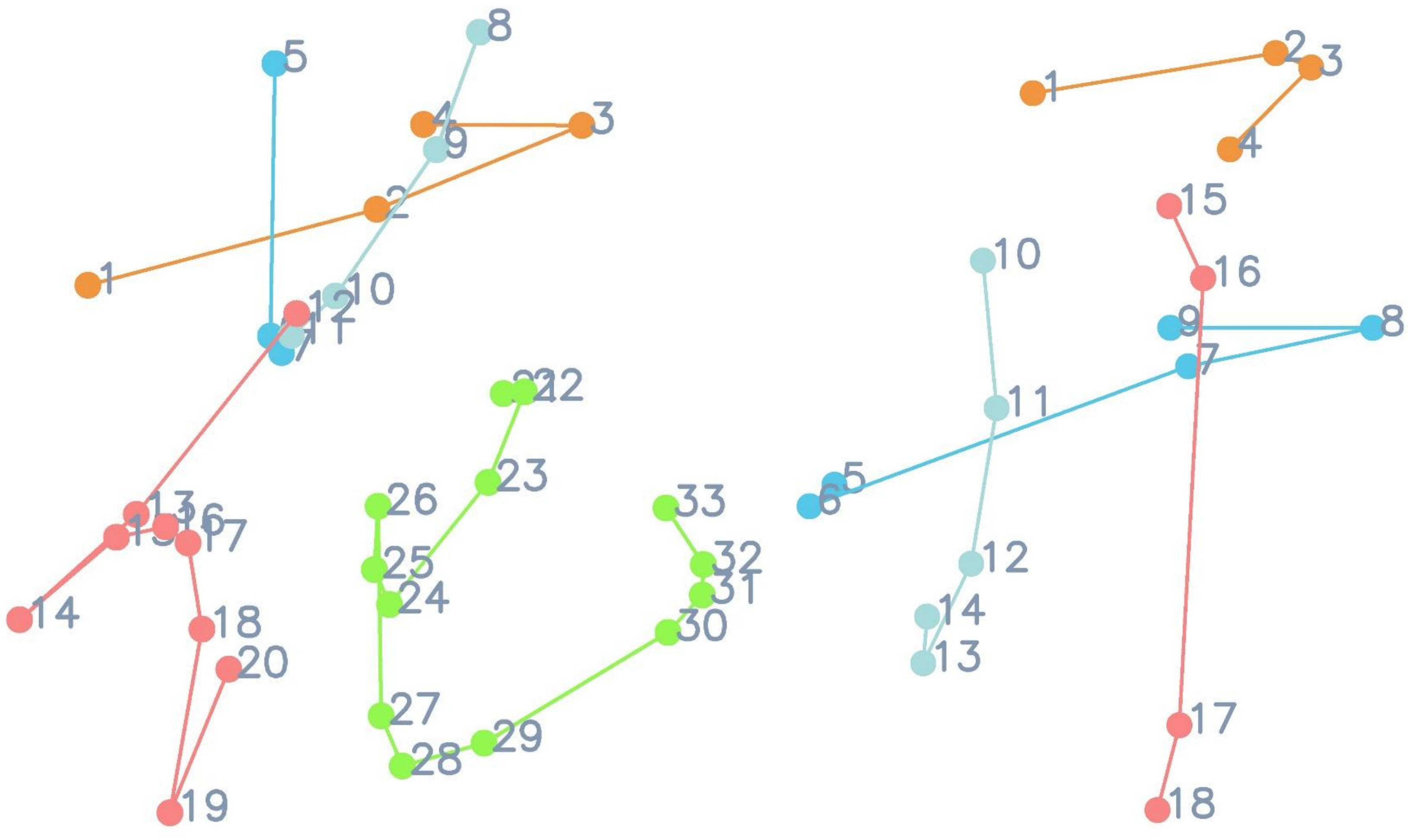}
\vspace{-0.05in}
\caption{Illustration of two online handwritten Chinese characters, with each color representing a stroke. The increasing numbers indicate the writing order from the start to the end.}
\label{fig:online}
\vspace{-0.2in}
\end{figure}

As the oldest writing system, Chinese characters are widely used across Asian countries. When compared with Latin scripts, Chinese characters encompass an exceptionally vast lexicon (87,887 characters in GB18030-2022 charset) and have intricate structures composed of multiple strokes. Recently, the  intriguing task of generating Chinese characters has garnered significant attention~\cite{zi2zi,gao2019artistic,liu2022xmp}. A promising approach to synthesising realistic handwritings is  to progressively generate online characters (\ie, the handwriting trajectory in a sequential format)~\cite{zhang2017drawing}. As shown in \cref{fig:online}, online characters convey richer information  (\eg, the order of writing)  and thus pave the way for various applications, including writing robots~\cite{yin2016synthesizing}.

\begin{figure}[t]\vspace{-0.1in}
\centering
\includegraphics[width=0.85\linewidth]{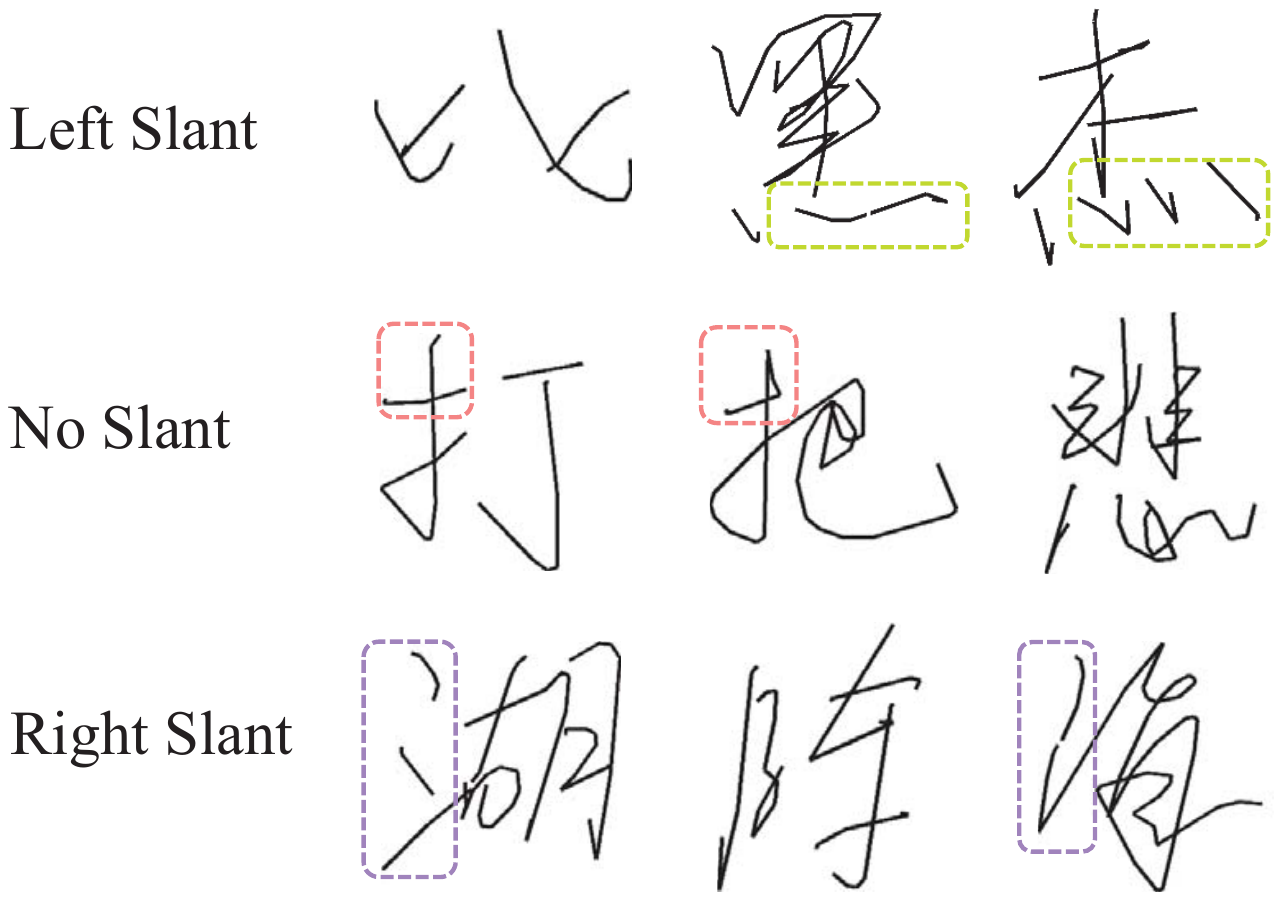}
\vspace{-0.1in}
\caption{Handwritten character samples from three unique writers, with each row containing characters by the same person. Despite sharing similar overall handwriting styles (\eg, glyph slant), subtle style differences (\eg, stroke length, location, and curvature) can still be observed among them.}
\label{fig:style_inconsisitens}
\vspace{-0.1in}
\end{figure}

Our goal is to automatically generate online Chinese handwritings that not only correspond to specific textual content, but also emulate the calligraphic style of a given exemplar writer (\eg, glyph slant, shape, stroke length, and curvature). This task thus holds potential for a wide range of applications, such as font design and calligraphy education.  A popular solution~\cite{kang2020ganwriting} is to extract style information from the provided stylized samples and merge it with the content reference.  DeepImitator~\cite{zhao2020deep} concatenates the style vector obtained from a CNN encoder with a character embedding, which is then fed into the RNN decoder to generate stylized online characters. WriteLikeYou~\cite{tang2021write} adopts the large-margin softmax loss~\cite{wang2018cosface} to promote discriminative learning of style features. However, these methods mainly focus on the overall writing style, thus overlooking the detailed style inconsistencies (\eg, the highlighted regions in \cref{fig:style_inconsisitens}) between characters produced by the same writer.

\begin{figure}[t]\vspace{-0.1in}
    \centering
    \includegraphics[width=0.79\linewidth]{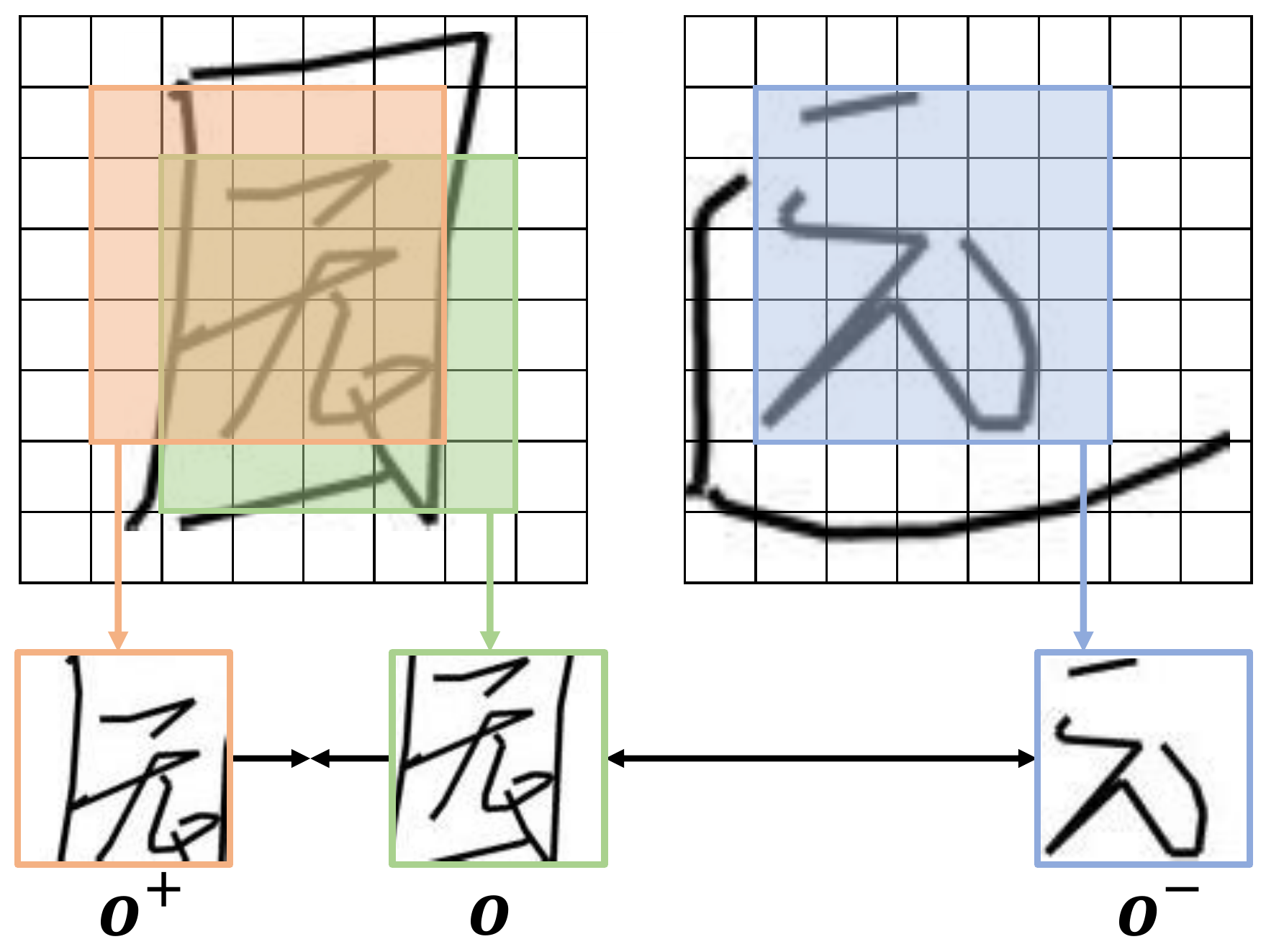}
    \vspace{-0.1in}
    \caption{In this two-character example, we independently sample the positive pair, \ie, $o$ and $o^+$, within the first character, while the negative $o^-$ is sampled from another character. Our sampling strategy randomly selects a small subset of patches, following a uniform distribution.}
    \label{fig:patch}
    \vspace{-0.1in}
\end{figure}

The observations mentioned above inspire us to disentangle style representations at the writer and character levels from the stylized handwritings. However, accurately capturing these two styles is a challenging task. To address this, we propose a style-disentangled Transformer (SDT) equipped with a dual-head style encoder. Specifically, we employ the contrastive learning framework~\cite{hadsell2006dimensionality} to guide each head in concentrating on writer-wise and character-wise styles, respectively. For the overall writer-wise style, we treat characters from the same writer as positive instances, and characters from different writers as negatives. This  enables the encoder to learn the style commonalities among characters written by the same writer. Regarding the detailed character-wise style, we independently sample positive pairs within a character, and sample negative samples from other characters, as illustrated in \cref{fig:patch}. Aggregating positive views of a character encourages the encoder to focus on the intricate character style patterns.

In addition, we introduce a content encoder for SDT to learn a textual feature with a global context. The two style representations, along with the textual feature, are then fed into a decoder that progressively generates online characters. Given that the output characters are in sequential form, we employ Transformer~\cite{vaswani2017attention}, a powerful sequence modeling architecture, as our backbone.
 
To extend SDT for generating offline Chinese handwrittings (\ie, character images with stroke-width, \eg,~\cref{all_offline,offline_1,offline_2,offline_3,offline_4} in Appendix), we further propose an offline-to-offline generation framework. We first use SDT to generate online characters with significant shape changes, and then decorate them with stroke width, ink-blot effects, \etc. This enables us to generate authentic offline handwritings. For more details, please refer to Appendix~\ref{a.4}.

We summarize our contributions in three key aspects:
\begin{itemize}
  \item We are the first to disentangle two style representations (\ie, writer-wise and character-wise) for enhancing Chinese handwriting generation. Our findings show that the former focuses more on low-frequency information, while the latter captures higher frequencies.
  \item We introduce a novel online character generation method, \emph{\ie,} SDT. 
  Extensive experiments on handwriting datasets in Chinese, English, Japanese, and Indic scripts demonstrate its effectiveness and superiority.
  \item Building on the SDT, we further develop an offline-to-offline framework that can produce more plausible offline handwritten Chinese characters, as evidenced in Appendix~\ref{a.4}.
 
\end{itemize}

\section{Related Work}

\noindent \textbf{Handwriting generation}.
Various style-content disentangling methods~\cite{aksan2018deepwriting, kang2020ganwriting,kotani2020generating,gan2021higan,luo2022slogan} have been proposed to generate handwritings with arbitrary styles. These methods assume that reference samples can be decomposed into style and content spaces. They disentangle calligraphic styles from reference samples and recombine them with specific textual content for controllable style synthesis. For instance,  DeepWriting~\cite{aksan2018deepwriting} adopts RNNs to extract the style vectors from online handwritings, and then combines them with character embeddings for synthesizing stylized online handwritings. However, it typically over-smooths the styles of distinct letters and loses key details, since its extracted style vectors are letter-independent~\cite{kotani2020generating}.

DSD~\cite{kotani2020generating} addresses the over-smooth problem by segmenting online handwritten words into isolated letters and encoding them into global and letter-specific style vectors, improving synthetic handwriting quality.  Compared with it, our method can explicitly capture more fine-grained details (\eg, stroke length, location and curvature), which are more difficult to obtain. Besides, these methods~\cite{aksan2018deepwriting,deepwritesyn,kotani2020generating} rely on  extra fine annotations to segment input cursive scripts, while our SDT need not. Recently, HWT~\cite{bhunia2021handwriting} uses a vanilla Transformer encoder for style pattern extraction. However, these methods~\cite{aksan2018deepwriting,kotani2020generating,bhunia2021handwriting,gan2021higan,kang2020ganwriting} rely on complex content references, such as recurrent embeddings and letter-wise filter maps. SLOGAN~\cite{luo2022slogan} addresses this issue by extracting textual contents from easily obtainable printed images, but  it struggles with generalizing to unseen handwriting styles  due to the fixed writer ID.  In contrast, our SDT effectively obtains content and style information, and thus can synthesize characters with arbitrary styles well. 

\begin{figure*}[t]
\vspace{-0.1in}
\centering
\includegraphics[width=0.75\linewidth]{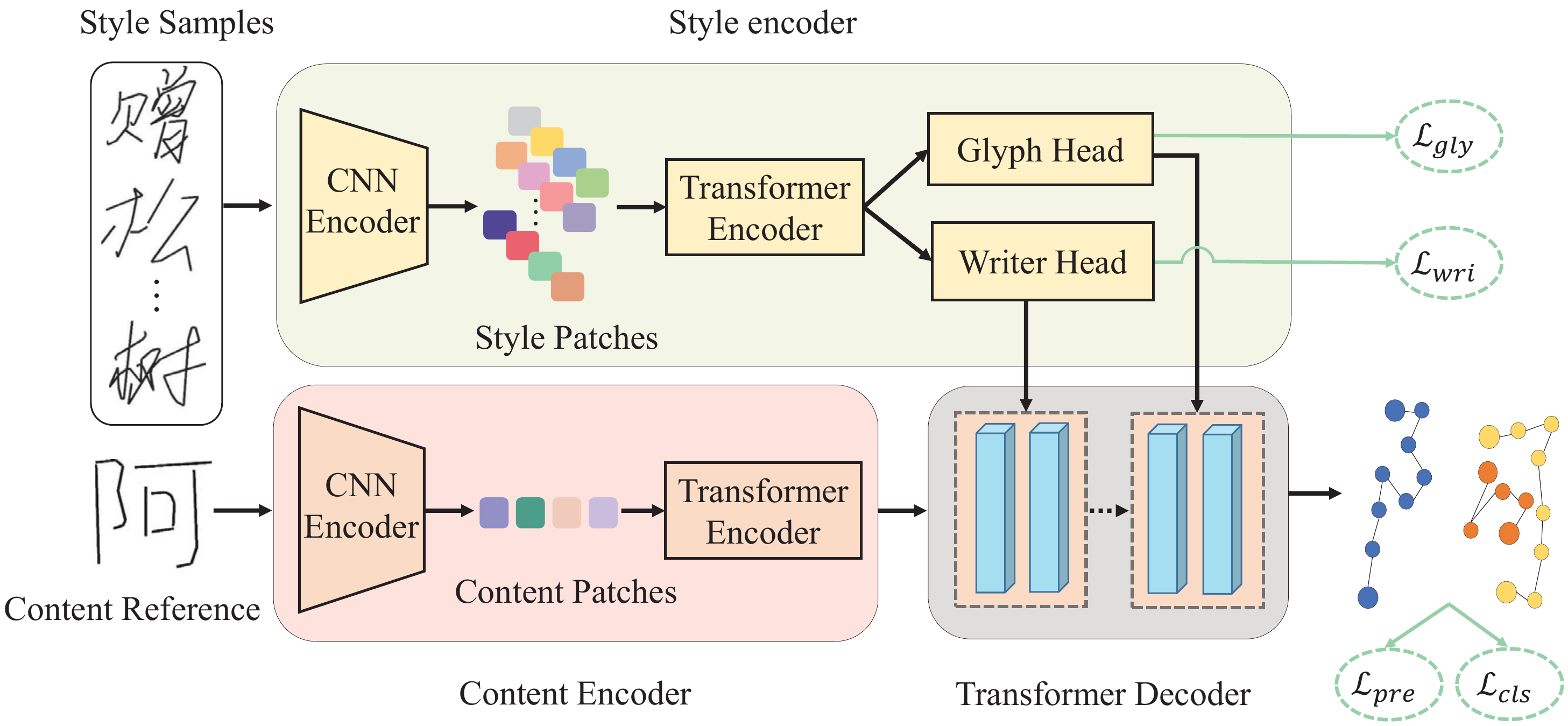}
\caption{Overview of the proposed method. 
Our SDT consists of a dual-head style encoder, a content encoder and a Transformer decoder. The writer-wise and character-wise style representations extracted by style encoder and the learned content feature are fed into the Transformer decoder to progressively generate online handwritings.  We utilize the WriterNCE $\mathcal{L}_{wri}$ and GlyphNCE $\mathcal{L}_{gly}$ to guide the two heads for learning the corresponding styles. $\mathcal{L}_{pre}$ and $\mathcal{L}_{cls}$ denote the pen moving prediction and state classification loss, respectively.} 
\label{fig:overview}
\vspace{-0.1in}
\end{figure*}

Regarding handwritten Chinese characters, several attempts~\cite{kong2017handwritten, chang2018generating,huang2022AGTGAN} use GANs to generate Chinese handwriting images, but always result in characteristic artifacts. Drawing~\cite{zhang2017drawing} and FontRNN~\cite{tang2019fontrnn} adopt RNNs,  but they cannot flexibly control the handwriting styles. In addition, DeepImitator~\cite{zhao2020deep} and WriteLikeYou~\cite{tang2021write} propose style-controlled generation using style representations from offline images~\cite{zhao2020deep} or online trajectories\cite{tang2021write}. Unlike these methods~\cite{zhao2020deep, tang2021write} that only extract an overall writer style, our SDT captures both writer and character-level styles, significantly improving  the performance of handwriting imitation.

~

\noindent \textbf{Contrastive learning}.
Contrastive learning~\cite{hadsell2006dimensionality,zhang2021unleashing}  has been widely used in various fields~\cite{TianKI20,GaoYC21,ren2022learning,qiu2021source,zhangself,zhang2021deep}. Some image translation studies~\cite{park2020contrastive, han2021dual} use contrastive learning to enhance natural image generation quality, encouraging content preservation during transfer. Unlike these methods~\cite{park2020contrastive, han2021dual}, our SDT aligns independently sampled patches from the same input, aiming to improve style representations of handwritings.

\section{Method}
\noindent \textbf{Problem statement.}
We aim to synthesize stylized online handwritings with conditional content and style. Given a content image $I$ and a set of style images $X_s=\{x_s^i\}_{i=1}^K$, randomly sampled from a writer $w_s$, we aim to generate an online handwritten Chinese character $\hat{Y}_s$ that reflects the calligraphic style of $w_s$ and retains the textual content of~$I$. The key challenge lies in obtaining discriminative style representations from a limited number of stylized samples.

To address this task, inspired by our observations (cf.~\cref{fig:style_inconsisitens}) of the \textit{overall uniformity} (\ie, writer-wise style) and \textit{inconsistent details} (\ie, character-wise style), we propose to disentangle the style of exemplar writers into overall and detailed styles for enhancing handwriting imitation performance. To achieve this, we introduce a new style-disentangled Transformer (SDT) approach to decouple the two styles from individual handwritings. The overall scheme of SDT is presented below.

\subsection{Overall Scheme}\label{overview} 
As shown in~\cref{fig:overview}, SDT consists of a dual-head style encoder, a content encoder, and a Transformer decoder. The style encoder (cf.~\cref{style_encoder}) is designed to learn the writer-wise and character-wise styles. It firstly extracts rich calligraphic style patterns from reference style examples $X_s$ via a sequential combination of a CNN and a Transformer encoder, followed by the writer and glyph heads to disentangle the writer-wise and character-wise styles from the extracted style patterns. To this end, we propose two contrastive objectives, WriterNCE $\mathcal{L}_{wri}$ and GlyphNCE $\mathcal{L}_{gly}$, to encourage the encoder to learn the two styles, respectively. Specifically,   $\mathcal{L}_{wri}$ maximizes the mutual information between character instances belonging to the same writer, while   $\mathcal{L}_{gly}$ associates the positive pair independently sampled from the same character. Guided by $\mathcal{L}_{wri}$ and $\mathcal{L}_{gly}$, the writer and glyph heads can acquire discriminative writer-wise and character-wise style features, respectively.

The content encoder uses a standard Resnet18~\cite{he2016deep} as the CNN backbone to learn the compact feature map $q_{map} \in \mathbb{R}^{h \times w \times c}$ from a content reference $I$, and feeds the flattened feature patches into a Transformer encoder to extract the textual content representation $q \in \mathbb{R}^{d \times c}$, where $d = h \times w$ and $c$ is the channel dimension. Benefiting from the strong capability of Transformer to capture long-range dependencies between feature patches, the content encoder expects an informative content feature $q$ with a global context. 

After the two encoders, a multi-layer Transformer decoder (cf. Section~\ref{Transformer_decoder}) is used to synthesize $\hat{Y}_s$ in an auto-regressive fashion, conditioned on the two style representations and the content feature. This decoder is supervised by the pen moving prediction loss $\mathcal{L}_{pre}$ and pen state classification loss $\mathcal{L}_{cls}$. 

To summarize, the overall training objective of SDT combines all four loss functions:
\begin{equation}\label{total}
\mathcal{L} = \mathcal{L}_{wri} + \mathcal{L}_{gly} + \mathcal{L}_{pre} + \lambda\mathcal{L}_{cls},
\end{equation}
where $\lambda$  serves as a trade-off factor. Each component of our SDT is detailed in the following \cref{style_encoder} and \cref{Transformer_decoder}.


\subsection{Dual-head Style Encoder}\label{style_encoder}
As illustrated in~\cref{fig:style_inconsisitens}, there are two distinct styles in a person's handwriting: writer-wise and character-wise styles. We propose a dual-head style encoder to obtain the two style representations. As shown in~\cref{fig:overview}, the input $X=\{x^i\}_{i=1}^K$ is firstly encoded by  ResNet18 to obtain a sequence of feature maps $F_m = \{f_m^i\}_{i=1}^K \in\mathbb{R}^{K \times h \times w \times c}$. Next, we flatten the spatial dimension of each feature map to obtain feature sequences $F = \{f^i\}_{i=1}^K \in\mathbb{R}^{K \times d \times c}$. These feature sequences are then fed into a Transformer encoder to extract more informative feature sequences $Z = \{z^i\}_{i=1}^K \in\mathbb{R}^{K \times d \times c}$. 
Finally, we use the two heads, built on  self-attention~\cite{vaswani2017attention},  to further disentangle $Z$ into the writer-wise style representations $E = \{e^i\}_{i=1}^K \in\mathbb{R}^{K \times d \times c}$ via $\mathcal{L}_{wri}$ and the character-wise counterparts $G = \{g^i\}_{i=1}^K \in\mathbb{R}^{K \times d \times c}$ via $\mathcal{L}_{gly}$, respectively. We next detail the two contrastive learning objectives as follows.

\subsubsection{Writer-wise Contrastive Learning}\label{writer_style}
To explicitly encourage the writer head to learn the writer-wise style, we propose to learn a feature space where  the style features from the same writer are closer than those from different writers. The intuition is that one person's handwritings consistently exhibit similar style information, which can serve as a crucial clue for distinguishing writers. To this end,  we take characters written by the same person as positive pairs and those from different writers as negative samples, and develop a new WriterNCE loss for writer-wise style learning.

Specifically, let $j \in M=\{1, ..., N\} $ be the index of an  element within a mini-batch and let $A\left(j\right)=M \backslash \{j\}$ be other indices distinct from $j$, where $N$ is the batch size. Given a writer-wise style feature $e_j$ belonging to writer $w_j$ as the anchor, we denote its in-batch positive sample set as $P(j)=\{p \in A\left(j\right):w_p=w_j\}$ and its negative set as $A\left(j\right) \backslash P\left(j\right)$. The WriterNCE loss is formulated as follows: 
\begin{equation}\label{WriterNCE}
\mathcal{L}_{wri}\small{=}\frac{\small{-}1}{N}\sum_{j \in M}\frac{1}{	\left|P\left(j\right)\right|}\sum_{p \in P\left(j\right)}\log \frac{\exp{\left({{\rm sim}\left(e_j,e_p\right)}/\tau\right)}}{\sum_{a \in A\left(j\right)}\exp{\left({{\rm sim}\left(e_j,e_a\right)}/\tau\right)}},
\end{equation}
where ${\rm sim}\left(e_j,e_p\right)\small{=}f_1\left(e_j\right)^{\top}f_1(e_p)$, $\tau$ is a temperature parameter and $f_1\left(\cdot\right)$ is a multi-layer perceptron (MLP) that projects features to a $\ell_2$-normalized feature space where $\mathcal{L}_{wri}$ is applied.  

\subsubsection{Character-wise Contrastive Learning}\label{character_style}
Compared with the overall writer-wise style,   character-wise style differences often exist in the fine details of distinct characters, \eg, stroke length and curvature (cf. ~\cref{fig:patch}). Inspired by this, we aim to maximize the mutual information between diverse views of a character, thereby enforcing the glyph head to learn the  character-wise style. As strokes are distributed across arbitrary spatial locations in character images, we propose to capture  stroke details via contrastive learning, by randomly selecting a small subset of patches following a uniform distribution.  Specifically, we conduct sampling over the sequential patch tokens obtained from the glyph head.

Given character-wise style features $\{g_j\}_{j=1}^{B}\in\mathbb{R}^{B \times d \times c}$ extracted from $B$ characters, we sample a positive patch pair (\ie, $o \in \mathbb{R}^{n \times c}$ and $o^+ \in \mathbb{R}^{n \times c}$) from the same randomly selected $g$, and $B\small{-}1$ negative patches $\{o_j^-\}_{j=1}^{B\small{-1}}$ from the remaining $B\small{-}1$ style features. Here, the number of sampled patch tokens is $n=d\cdot\alpha$, where $\alpha$ is the sampling ratio. The GlyphNCE loss is formulated as:
\begin{equation}\label{GlyphNCE}
\mathcal{L}_{gly}\small{=-}\log \frac{\exp{\left({{\rm sim}\left(o,o^+\right)}/\tau\right)}}{\exp{\left({\rm sim}{\left(o,o^+\right)}/\tau\right)}\small{+}\sum_{j\small{=}1}^{B\small{-}1}\exp{\left({\rm sim}{\left(o,o_j^-\right)}/\tau\right)}},
\end{equation}
where ${\rm sim}\left(o,o^+\right)\small{=}f_2\left(o\right)^{\top}f_2(o^+)$, and $f_2(\cdot)$ is an MLP with the same structure as $f_1(\cdot)$.

\begin{figure}[t]
\centering  
\includegraphics[width=0.68\linewidth]{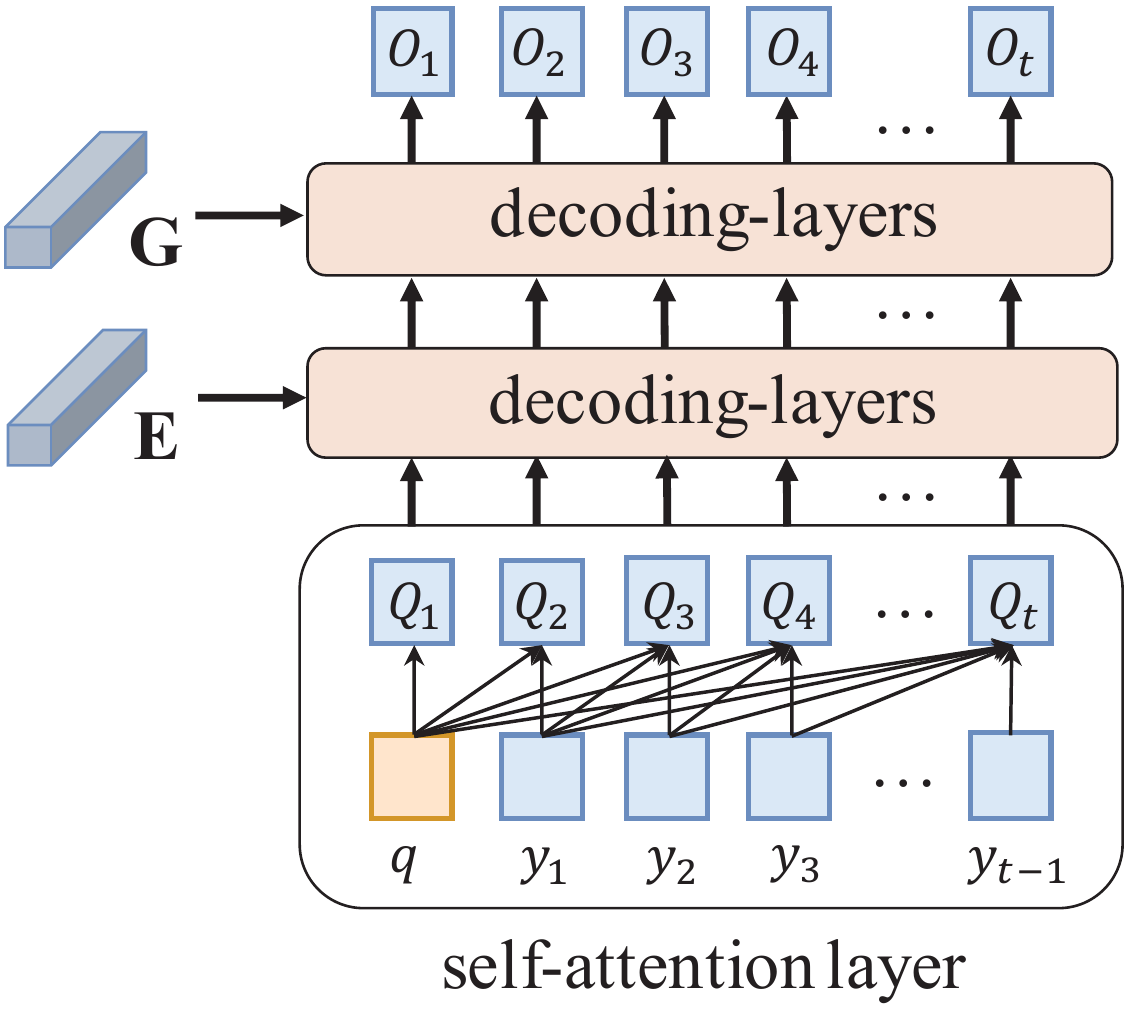}
    \caption{Illustration of the fusion of the style information in our Transformer decoder.  At each time step, the query vector is first encoded from the content feature $q$ and previous points $\{y_j\}_{j=1}^{t-1}$. Then, it successively attends to the writer-wise and character-wise style features (\ie, $E$ and $G$) for adaptively aggregating style information, which is finally decoded into the current-step output.}
    \label{fig:decoder}
    \vspace{-0.1in}
\end{figure}

\subsection{Transformer Decoder for Handwriting}\label{Transformer_decoder}
The goal of the Transformer decoder is to progressively generate realistic online characters, denoted as $\hat{Y}$,  based on the global content feature $q$ and the obtained style representations (\ie, $E = \{e^i\}_{i=1}^K$ and $G = \{g^i\}_{i=1}^K$). As each online character consists of numerous points (\ie, $\hat{Y}=\{{\hat{y}_j}\}_{j=1}^{L}$, with $L$ being the total number of points; see more details in Appendix~\ref{a.10}),  the decoder faces the challenge of effectively integrating content and style features to accurately depict all points of the character. To address this challenge,  we propose to fuse $q$, $E$ and $H$ within the multi-head attention layers of our Transformer decoder. As shown in \cref{fig:decoder},  $E$ and $H$ serve as the key/value vectors, while  $Q$ serves as the query vector that successively attends to $E$ and $G$ to aggregate style information.   
  
During training, at any decoding step $t$, we take $q$ as the initial point. As shown in~\cref{fig:decoder}, we apply self-attention 
to the point sequence, \ie the content context $\left[q, y_1, ..., y_{t-1}\right]$, to obtain the query vector $Q_t \in \mathbb{R}^c$. Here, the ground-truth points $\{{y_j}\}_{j=1}^{t-1}$ are used to accelerate model convergence in training~\cite{williams1989learning,tang2021write}. Subsequently, $Q_t$ attends to $E$ and $G$ over the subsequent decoding layers to adaptively aggregate style information, ultimately generating the output $O_t \in \mathbb{R}^{6R+3}$. The output with $6R+3$ logits is then used to generate the pen moving $\left(\Delta{\hat{u}_t}, \Delta{\hat{v}_t}\right)$ and  the pen state $\left( \hat{m}_t^1, \hat{m}_t^2, \hat{m}_t^3 \right)$. Specifically,  we use a Gaussian mixture model (GMM)~\cite{graves2013generating} with $R$ bivariate normal distributions to predict the pen moving, with each normal distribution containing 6 parameters. Moreover, we use the other $3$  logits  to generate the pen state. The training loss for supervising the decoder comprises two parts: the pen movement prediction loss  $\mathcal{L}_{pre}\left(\Delta{\hat{u}_t}, \Delta{\hat{v}_t}\right)$, and the pen state classification loss $\mathcal{L}{cls}\left(\hat{m}_t^1, \hat{m}_t^2, \hat{m}_t^3\right)$.  Please refer to Appendix~\ref{a.11} for more details of these losses.

The inference phase is different from the training phase, where the ground truth $y$ is not available at test time. Instead, we take the generated points$\{{\hat{y}_j}\}_{j=1}^{t-1}$ as the input of the step $t$, and combine them with $q$, $E$, and $G$ to predict the next point $\hat{y}_t$. Such a  process repeats until a pen-end state ($\hat{m}_{t-1}^3\small{=}1$) is received. 

\section{Experiments}
\subsection{Chinese handwriting generation}

\noindent \textbf{Dataset.}
To evaluate  SDT in generating Chinese handwritings, we use CASIA-OLHWDB~(1.0-1.2)~\cite{liu2011casia} for model training and ICDAR-2013 competition database~\cite{yin2013icdar} for testing. The training set consists of about 3.7 million online Chinese handwritten characters produced by 1,020 writers, while the test set contains 60 writers, with each contributing 3,755 most frequently used characters set of GB2312-80. Following~\cite{ICLRSKETCH}, we use the Ramer–Douglas–Peucker algorithm~\cite{douglas1973algorithms} with a parameter of $\epsilon=2$ to remove redundant points of characters, leading to an average sequence length of 50. Following ~\cite{zhao2020deep}, we render offline style references using coordinate points of online characters, and each style sample is randomly sampled from the target-writer handwritings during inference, as shown in~\cref{fig:overview}. For content images, we use the popular average Chinese font~\cite{jiang2019scfont}.

\begin{figure*}[t]
    \centering
    \includegraphics[width=0.82\linewidth]{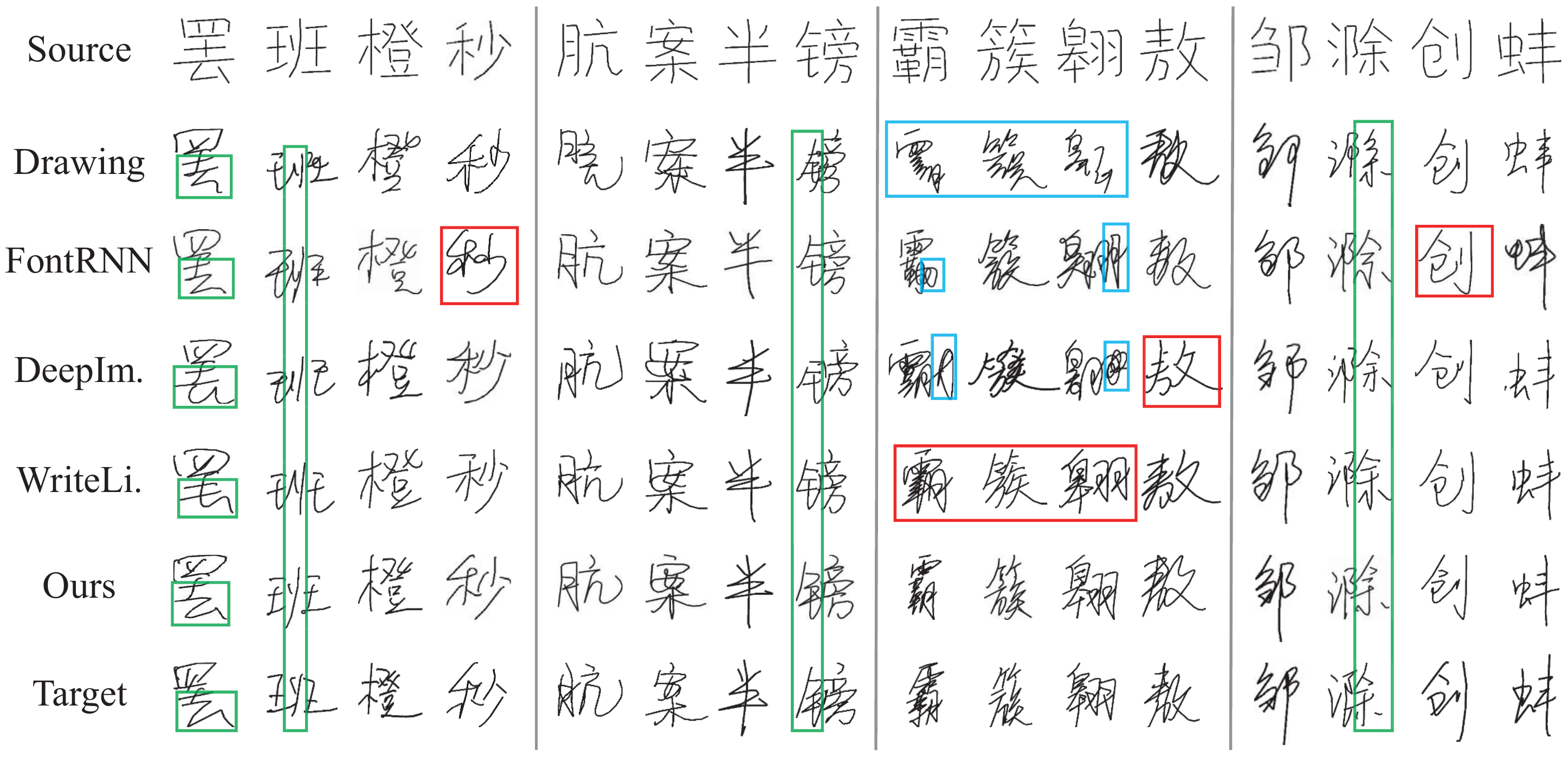}
    \vspace{-0.1in}
    \caption{Comparisons with the state-of-the-art methods for online Chinese handwriting generation. The red and blue boxes highlight failures of style imitation and structure preservation, respectively. WriteLi. indicates the WriteLikeYou-v2. The green boxes highlight comparisons between the fine details of targets and generated characters.}
    \label{fig:sota_method}
    \vspace{-0.1in}
\end{figure*}

\noindent \textbf{Evaluation metrics}\label{QEM}.
We use Dynamic Time Warping (DTW)~\cite{berndt1994using,chen2022complex}, an elastic matching technique for aligning the given two sequences, to calculate the distance between the generated and real characters. Moreover, we use Content Score ~\cite{zhao2020deep} to measure the structure correctness of generated characters, and adopt Style Score~\cite{tang2021write} to quantify the style similarity between the generated and real handwritings.
We also conduct user preference studies to quantify the subjective quality of the output characters. More details are provided in Appendix~\ref{a.1.1}.

\noindent\textbf{Implementation details}. We set the number of the style reference to $K=15$, and resize the reference style and content images to $64 \times 64$. Moreover, each Transformer encoder contains 2 self-attention layers, while the Transformer decoder has 4 layers for obtaining style representations (2 for writer-wise and 2 for character-wise).  Following the original Transformer~\cite{vaswani2017attention}, each Transformer layer contains the multi-head attention with $c= 512$ dimensional states and 8 attention heads.  Contrary to HWT~\cite{bhunia2021handwriting}, where $F$ is concatenated before the Transformer encoder, we process each feature sequence $f \in F$ individually.  Moreover, we apply sinusoidal positional encoding~\cite{vaswani2017attention} to  input tokens before feeding them to the Transformer encoder and decoder. For training, we first pre-train the content encoder with 138k iterations (batch size of 256) for character classification on training samples and then train the whole model with 148k iterations (batch size of 128), on a single RTX3090 GPU. The optimizer is Adam~\cite{kingma2015adam,zhuang2021new}, with a learning rate of 0.0002 and gradient clipping of 5.0. The sampling ratio $\alpha$ is determined through a search over ${0.25, 0.5, 0.75, 1.00}$, and 0.25 is chosen. Following ~\cite{tang2021write}, we set $\lambda=2$. Further implementation details are provided in Appendix~\ref{a.1.2}.


\noindent \textbf{Compared methods}. We compare our proposed SDT with state-of-the-art online Chinese character generation methods, \ie Drawing~\cite{zhang2017drawing}, FontRNN~\cite{tang2019fontrnn}, DeepImitator~\cite{zhao2020deep}, and WriteLikeYou~\cite{tang2021write}. To ensure a fair comparison, we re-implement Drawing and FontRNN by adding the style branch of DeepImitator~\cite{zhao2020deep}, enabling them to achieve arbitrary stylized character generation. To adapt WriteLikeYou~\cite{tang2021write} to handle images, we update its encoder to create two new variants: WriteLikeYou-v1 (\ie, CNN encoder~\cite{zhao2020deep}) and WriteLikeYou-v2 (\ie, the same CNN-Transformer encoder as our SDT), based on the released official code\footnote{\url{https://github.com/ShusenTang/WriteLikeYou}}. More  details are available in Appendix \ref{a.1.3}.

\begin{figure*}[t]
    \centering
    \includegraphics[width=0.9\linewidth]{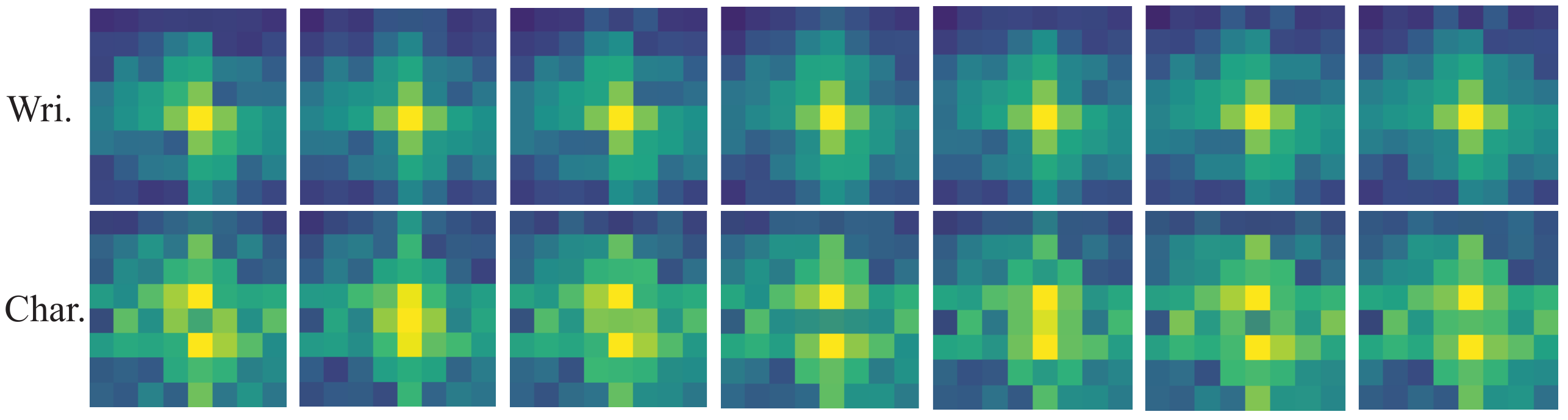} 
    \vspace{-0.1in}
    \caption{Spectrum analysis of two style representations. We provide frequency magnitude visualizations belonging to $7$ writers, where the top row shows the writer-wise style while the bottom represents the character-wise one. Each spectrum map is averaged over 100 character samples written by the same person. The brighter the color, the larger the magnitude. A pixel farther from the center means a higher frequency~\cite{pan2022hilo}. We find that the writer-wise style representations capture more low-frequency information, while character-wise style representations capture more high-frequency information.}
    \vspace{-0.1in}
    \label{fig:fft}
\end{figure*}

\begin{table}
	\caption{Comparisons with  state-of-the-art methods on  Chinese dataset. See more results of WriteLikeYou~\cite{tang2021write} in Appendix~\ref{a.6}.}
    \vspace{-0.2in}
    \begin{center}
    \scalebox{0.8}{
    \begin{threeparttable} 
	\begin{tabular}{lcccc}\toprule
        Method  &  \makecell[c]{Style \\ Score $\uparrow$}  & \makecell[c]{Content \\ Score $\uparrow$} & DTW $\downarrow$ & \makecell[c]{User \\ Prefer. $\left(\%\right)$ $\uparrow$} \\ \midrule 
        Drawing~\cite{zhang2017drawing} & 35.83 &78.15 & 1.1813 & 3.53  \\
        FontRNN~\cite{tang2019fontrnn} & 46.14 & 92.18 & 1.0448 & 7.07 \\
        DeepImitator~\cite{zhao2020deep} & 50.67 &90.92 & 1.0622 & 7.99  \\
        WriteLikeYou-v1~\cite{tang2021write} & 71.09 &93.89 & 0.9832 & 11.67  \\
        WriteLikeYou-v2~\cite{tang2021write} & 72.37 &96.44 & 0.9289 & 13.07 \\
        SDT(Ours) & \textbf{94.50} & \textbf{97.04} & \textbf{0.8789} & \textbf{56.67}  \\
        \bottomrule
	\end{tabular} 
    \end{threeparttable}
    }
    \end{center}
    \label{tab:com_sota} 
    \vspace{-0.1in}
\end{table}

\noindent \textbf{Quantitative comparison}. Quantitative results are  in~\cref{tab:com_sota}, revealing that SDT outperforms other methods across all evaluation metrics. Notably, SDT surpasses the second-best method by a significant 22.13\% margin in Style Score, demonstrating the proposed method's exceptional style imitation performance.

\noindent \textbf{Qualitative comparison}. 
We visualize the generated samples of each method in~\cref{fig:sota_method}, which intuitively explains the significant superiority of SDT in the user preference study. In~\cref{fig:sota_method}, Drawing~\cite{zhang2017drawing} generates the least satisfactory results, often producing unreadable characters. FontRNN~\cite{tang2019fontrnn} and DeepImitator~\cite{zhao2020deep} occasionally synthesize unpleasant stroke paddings, and WriteLikeYou~\cite{tang2021write}  struggles with generating complex characters regarding style mimicry. In contrast, our method yields higher-quality results, particularly in recovering fine character details.
 
\subsection{Analysis}\label{abl}

In this section, we assess the impact of the two learned style features and that of the style inputs. We also analyze the effect of the sampling ratio $\alpha$, and discuss the combination strategies in the Transformer decoder in Appendix~\ref{a.5}. Besides, we discuss failure cases in Appendix~\ref{a.7}.
 
\noindent \textbf{Quantitative evaluation of two style representations.} We ablate  the effects of the two extracted style representations in~\cref{two_style}. The  findings include: (1) Both style representations enhance the quality of generated outcomes in all evaluation metrics, particularly in terms of Style Score, improving by 4.79\% and 5.86\%, respectively. (2) Combining the two style features further boosts model performance across all evaluation metrics, suggesting that the extracted styles are complementary. Moreover, the order in which the two style features are input into the Transformer decoder has no significant impact on  Style Score (93.72\% vs. 94.50\%).

\begingroup
\begingroup
\begin{table}[t!]
    \centering
    \caption{Ablation study. Here, we further use FID to measure the distance between generated and real samples for each writer separately, and finally average them.} 
 
    \label{two_style}
    \renewcommand\arraystretch{2.15}{
    \scalebox{0.6}{
    \begin{tabular}{cc |ccc |ccc}
     \hline
     writer-wise & character-wise & \multicolumn{3}{c|}{Generated Samples} & Style Score$\uparrow$ & FID$\downarrow$  & DTW$\downarrow$  \\
        \hline
         &  &    \includegraphics[scale=0.35,valign=c]{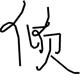} &
        \includegraphics[scale=0.35,valign=c]{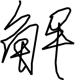} &
        \includegraphics[scale=0.35,valign=c]{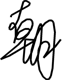}  & 85.52  & 27.75& 0.8941  \\
        \checkmark &  & \includegraphics[scale=0.35,valign=c]{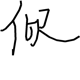} &
        \includegraphics[scale=0.35,valign=c]{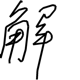} &
        \includegraphics[scale=0.35,valign=c]{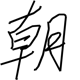} & 91.38 & 26.38&0.8841 \\
        & \checkmark &\includegraphics[scale=0.35,valign=c]{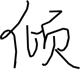} &
        \includegraphics[scale=0.35,valign=c]{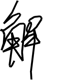} &
        \includegraphics[scale=0.35,valign=c]{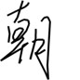} & 90.31&26.89&0.8803
        \\
        \checkmark & \checkmark & 
        
        \includegraphics[scale=0.35,valign=c]{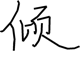} &
        \includegraphics[scale=0.35,valign=c]{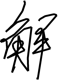} &
        \includegraphics[scale=0.35,valign=c]{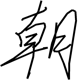} & 94.50&25.46&0.8789 \\
        \hline
        \multicolumn{2}{c|}{Ground Truth} &\includegraphics[scale=0.35,valign=c]{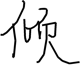} &
        \includegraphics[scale=0.35,valign=c]{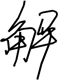} &
        \includegraphics[scale=0.35,valign=c]{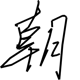}  &&&  \\
        \hline
    \end{tabular}}
    }
    \vspace{-0.1in}
\end{table}
\endgroup

 
\noindent \textbf{Qualitative comparison between two style representations.}
To further analyze the differences between the two styles, we resize the output patch tokens representing the two styles to feature maps, respectively, and visualize their frequency magnitudes in~\cref{fig:fft}. We find that character-wise style features capture more high-frequency information, whereas writer-wise features mainly focus on low-frequency information. According to~\cite{cooley1969fast}, the high-frequency information in an image usually captures fine details while the low frequencies contain the overall part of objects. This finding supports our motivation that the writer head helps to imitate the overall style (\eg, glyph slant), while the glyph head captures the detailed style (\eg, stroke curvature), as shown in~\cref{two_style}.


 \begin{figure}[t] 
    \centering
\includegraphics[width=0.82\linewidth]{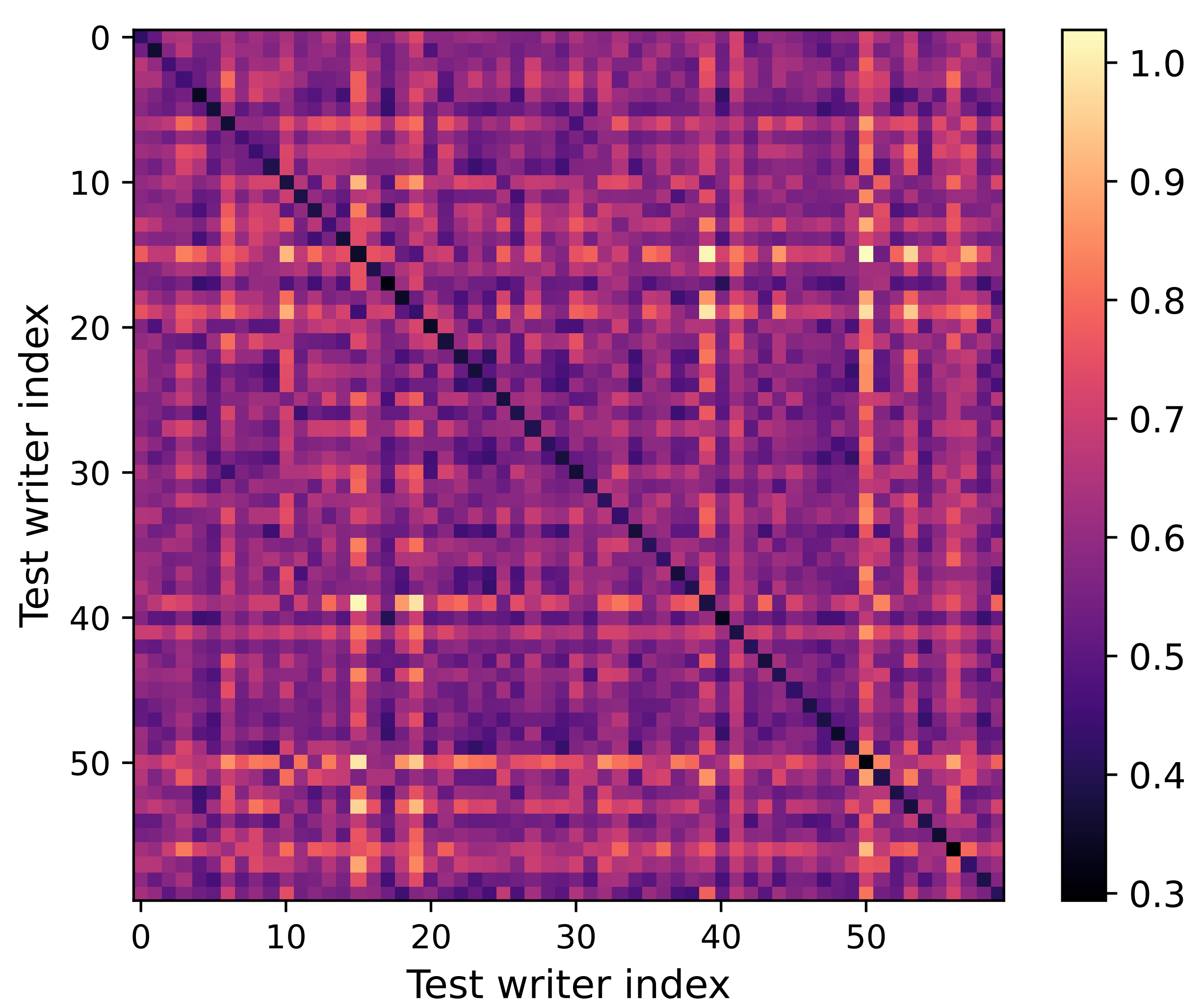}
    \vspace{-0.1in}
    \captionof{figure}{The heat map of the DTW matrix. The dark diagonal indicates that the generated characters still own a higher similarity even using different $X_s$ belonging to the same writer.}
    \vspace{-0.1in}
    \label{fig:dtw} 
\end{figure}

\begin{table}[t]
	\caption{Quantitative evaluations of our SDT and competitors on Japanese, Indic, and English datasets.}
    \vspace{-0.1in}
	\label{tab:foreign}
    \begin{center}
    \scalebox{0.85}{
    \begin{threeparttable} 
	\begin{tabular}{cccc}\toprule
    Datasets & Methods & Content Score$\uparrow$ & DTW $\downarrow$\\ \midrule 
     \multirow{4}{*}{Japanese} & Drawing~\cite{zhang2017drawing} & 50.74 & 1.4657  \\
     & DeepImitator~\cite{zhao2020deep} & 53.20 & 1.2564  \\
     &  WriteLikeYou-v2~\cite{tang2021write}   & 85.61 & 1.2066 \\
     &  SDT(Ours)  & \textbf{91.31} & \textbf{1.1289} \\
     \midrule 
     \multirow{4}{*}{Indic} & Drawing~\cite{zhang2017drawing} & 2.34 & 9.8230  \\
     & DeepImitator~\cite{zhao2020deep} & 4.13 & 6.7421  \\
     &  WriteLikeYou-v2~\cite{tang2021write}   & 13.19 & 4.5130 \\
     &  SDT(Ours)  & \textbf{97.22} & \textbf{0.7075} \\
     \midrule 
     \multirow{4}{*}{English} & Drawing~\cite{zhang2017drawing} & 79.14 & 1.8519  \\
     & DeepImitator~\cite{zhao2020deep} & 76.53 & 1.6460  \\
     &  WriteLikeYou-v2~\cite{tang2021write}   & 84.41 & 1.6215 \\
     &  SDT(Ours)  & \textbf{85.52} & \textbf{1.6048} \\
        \bottomrule
	\end{tabular} 
    \vspace{-0.1in}
    \end{threeparttable}}
    \end{center}
\end{table}

\begin{figure*}[t]
    \centering
    \includegraphics[width=0.9\linewidth]{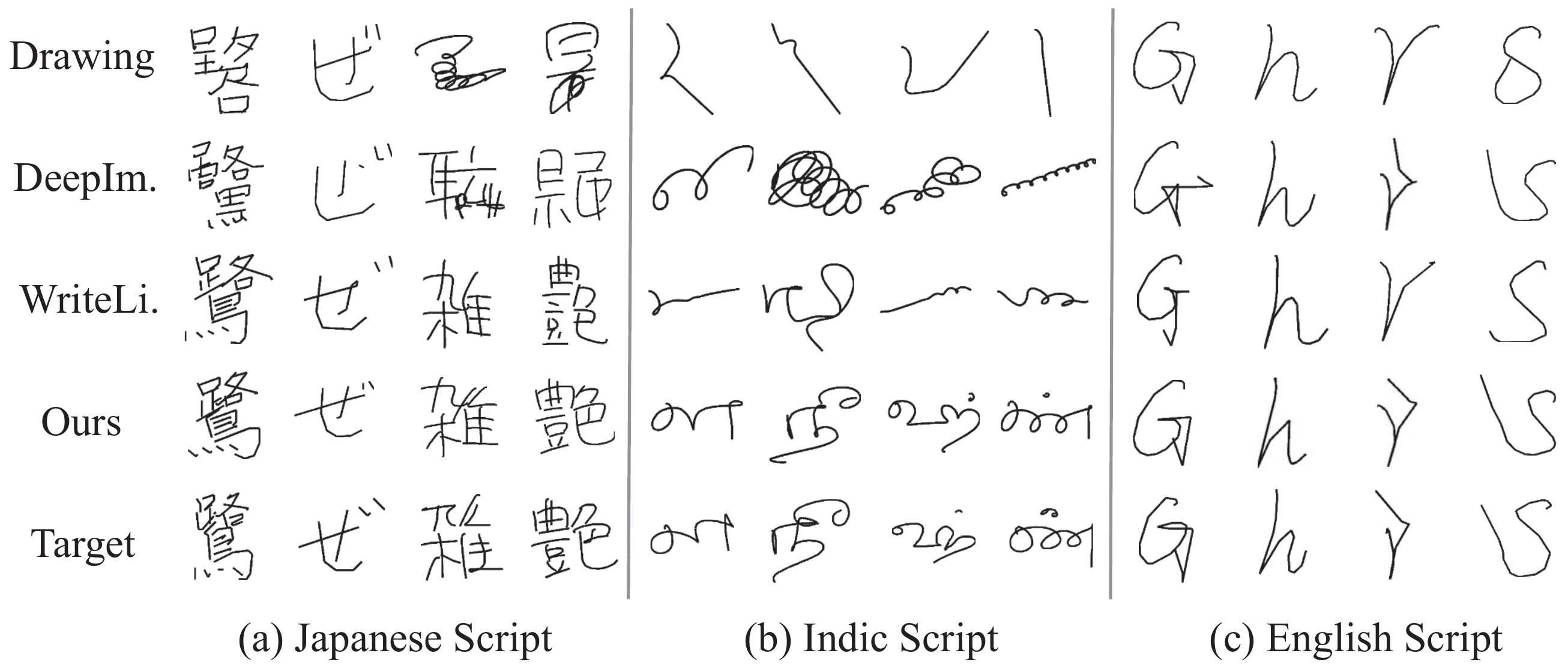} 
    \caption{Comparisons with the competitors for online handwriting generation on various scripts. WriteLi. indicates WriteLikeYou-v2.}
    \label{fig:foreign_com} 
\end{figure*}

\noindent \textbf{Effect of using different style inputs.} As mentioned in ~\cite{tang2021write}, given different style inputs $X_s$ belonging to the same writer $w_s$, the imitation model may generate inconsistent characters. To evaluate the effect of different style inputs, we conduct two independent experiments using different $X_s$ based on the same model. Following~\cite{tang2021write}, we use the model to generate 200 characters for each writer in the test set. We then calculate the DTW distance between corresponding characters individually and average them according to the writer index (see Appendix~\ref{a.1.4} for more details) to get a DTW square matrix. The resulting DTW square matrix is visualized in Figure~\ref{fig:dtw}. The dark diagonal in Figure~\ref{fig:dtw} suggests that the generated characters maintain a high degree of similarity even when using different $X_s$ belonging to the same writer, demonstrating that our SDT can generate consistent results from various style inputs.


\subsection{Applications to Other Languages}

\noindent \textbf{Japanese handwriting generation}.
For the Japanese handwriting generation task, we conduct experiments on TUAT HANDS~\cite{matsumoto2001collection} database~(see more details in Appendix \ref{a.1.5}) to evaluate the effectiveness of our method. \cref{fig:foreign_com} (a) and \cref{tab:foreign} verify the effectiveness of SDT for Japanese handwriting generation. Specifically, from \cref{tab:foreign}, we observe that our SDT outperforms all compared methods in terms of two quantitative metrics, indicating that  SDT performs well in multiple languages. Furthermore, we provide additional evaluation metrics in Appendix~\ref{a.8}.

\noindent \textbf{Indic handwriting generation}.
We evaluate our method on the Indic handwriting generation task based on the Tamil\footnote{\url{http://lipitk.sourceforge.net/datasets/tamilchardata.htm}} dataset.  It is worth noting that Indic handwriting generation presents a more significant challenge, as Indic characters contain more trajectory points than Chinese, Japanese, and English scripts (i.e., 88 vs. 68, 50, 30 on average; see Appendix~\ref{a.1.5} for more dataset information). We compare our method with other approaches on the official test set in terms of Content Score and Dynamic Time Warping (DTW). As shown in Table~\ref{tab:foreign},  we find that our SDT significantly outperforms the second-best method in terms of the two quantitative metrics, achieving an 84.03\% higher Content Score and a 3.8055 lower DTW. This indicates that our SDT can handle handwritten characters with a large number of points (averaging 88) and ensure the quality of synthetic samples, as illustrated in Figure~\ref{fig:foreign_com} (b). The potential advantages of our SDT are: (1) The improved style representations extracted by our SDT prevent the collapse of generated characters. (2) Our Transformer decoder facilitates long-distance dependence between trajectory points. We provide more experimental analysis in Appendix~\ref{a.9}.

 ~

\noindent \textbf{English handwriting generation}.
To evaluate our method in generating  English handwritings,  we collect all of the English samples from the symbol part of CASIA-OLHWDB(1.0-1.2)~\cite{liu2011casia} and ICDAR-2013 competition database~\cite{yin2013icdar} (see more details in Appendix~\ref{a.1.5}). Similarly, we use Content Score and DTW as evaluation metrics. As shown in Figure~\ref{fig:foreign_com} (c), we find that all methods achieve sound and comparable performance. One reason for this is that the English script contains fewer character classes and a smaller number of trajectory points (averaging 30), making their imitation easier compared to other scripts. Nevertheless, our SDT still outperforms other methods by a small margin in terms of both Content Score and DTW, as shown in Table~\ref{tab:foreign}. Moreover, we observe that corresponding uppercase and lowercase letters sometimes exhibit subtle inter-class differences (\eg, O vs. o), which leads our SDT to achieve a relatively low Content Score.

\section{Conclusion}
In this paper,  we have proposed a novel method, named style-disentangled Transformer (SDT), to synthesize realistic and diverse online handwritings.
SDT enhances imitation performance by disentangling the  writer-wise and  character-wise style representations from  individual handwriting samples. For the writer-wise style, we group characters from the same writer and separate those from different writers, promoting SDT's ability to learn uniformity in individual handwritings.  For the character-wise style, we  maximize the mutual information between the distinct views of a character. Moreover, we extend SDT and introduce an offline-to-offline framework for improving the generation quality of offline Chinese handwritings. Promising results on various language scripts verify the effectiveness of our SDT. Although primarily designed for handwriting generation, SDT still holds the potential for extension to other generative tasks, such as font generation.

\paragraph{Acknowledgments}
This research is partially supported by NSFC (Grant No.62176093, 61673182), Key Realm R\&D Program of Guangzhou (No.202206030001), Guangdong Basic and Applied Basic Research Foundation (No.2021A1515012282) and Guangdong Provincial Key Laboratory of Human Digital Twin (2022B1212010004).

\clearpage
{ 
\small
\balance
\bibliographystyle{ieee_fullname}
\bibliography{egbib}

\begin{thebibliography}{10}\itemsep=-1pt

\bibitem{aksan2020cose}
Emre Aksan, Thomas Deselaers, Andrea Tagliasacchi, and Otmar Hilliges.
\newblock Cose: Compositional stroke embeddings.
\newblock In {\em Advances in Neural Information Processing Systems}, pages
  10041--10052, 2020.

\bibitem{aksan2018deepwriting}
Emre Aksan, Fabrizio Pece, and Otmar Hilliges.
\newblock Deepwriting: Making digital ink editable via deep generative
  modeling.
\newblock In {\em Proceedings of the 2018 CHI Conference on Human Factors in
  Computing Systems}, pages 1--14, 2018.

\bibitem{MC-GAN}
Samaneh Azadi, Matthew Fisher, Vladimir~G Kim, Zhaowen Wang, Eli Shechtman, and
  Trevor Darrell.
\newblock Multi-content gan for few-shot font style transfer.
\newblock In {\em Computer Vision and Pattern Recognition}, pages 7564--7573,
  2018.

\bibitem{bengio1994learning}
Yoshua Bengio, Patrice Simard, and Paolo Frasconi.
\newblock Learning long-term dependencies with gradient descent is difficult.
\newblock {\em IEEE Transactions on Neural Networks}, pages 157--166, 1994.

\bibitem{berndt1994using}
Donald~J Berndt and James Clifford.
\newblock Using dynamic time warping to find patterns in time series.
\newblock In {\em ACM SIGKDD International Conference on Knowledge Discovery
  and Data Mining Workshop}, 1994.

\bibitem{bhunia2021handwriting}
Ankan~Kumar Bhunia, Salman Khan, Hisham Cholakkal, Rao~Muhammad Anwer,
  Fahad~Shahbaz Khan, and Mubarak Shah.
\newblock Handwriting transformers.
\newblock In {\em International Conference on Computer Vision}, pages
  1086--1094, 2021.

\bibitem{chang2018generating}
Bo Chang, Qiong Zhang, Shenyi Pan, and Lili Meng.
\newblock Generating handwritten chinese characters using cyclegan.
\newblock In {\em Winter Conference on Applications of Computer Vision}, pages
  199--207, 2018.

\bibitem{chen2022complex}
Zhounan Chen, Daihui Yang, Jinglin Liang, Xinwu Liu, Yuyi Wang, Zhenghua Peng,
  and Shuangping Huang.
\newblock Complex handwriting trajectory recovery: Evaluation metrics and
  algorithm.
\newblock In {\em Asian Conference on Computer Vision}, pages 1060--1076, 2022.

\bibitem{cooley1969fast}
James~W Cooley, Peter~AW Lewis, and Peter~D Welch.
\newblock The fast fourier transform and its applications.
\newblock {\em IEEE Transactions on Education}, 12(1):27--34, 1969.

\bibitem{douglas1973algorithms}
David~H Douglas and Thomas~K Peucker.
\newblock Algorithms for the reduction of the number of points required to
  represent a digitized line or its caricature.
\newblock {\em Cartographica: the international journal for geographic
  information and geovisualization}, pages 112--122, 1973.

\bibitem{fogel2020scrabblegan}
Sharon Fogel, Hadar Averbuch-Elor, Sarel Cohen, Shai Mazor, and Roee Litman.
\newblock Scrabblegan: Semi-supervised varying length handwritten text
  generation.
\newblock In {\em Computer Vision and Pattern Recognition}, pages 4324--4333,
  2020.

\bibitem{gan2021higan}
Ji Gan and Weiqiang Wang.
\newblock Higan: Handwriting imitation conditioned on arbitrary-length texts
  and disentangled styles.
\newblock In {\em AAAI Conference on Artificial Intelligence}, pages
  7484--7492, 2021.

\bibitem{GaoYC21}
Tianyu Gao, Xingcheng Yao, and Danqi Chen.
\newblock Simcse: Simple contrastive learning of sentence embeddings.
\newblock In {\em Empirical Methods in Natural Language Processing}, pages
  6894--6910, 2021.

\bibitem{gao2019artistic}
Yue Gao, Yuan Guo, Zhouhui Lian, Yingmin Tang, and Jianguo Xiao.
\newblock Artistic glyph image synthesis via one-stage few-shot learning.
\newblock {\em ACM Transactions on Graphics}, 38(6):1--12, 2019.

\bibitem{goodfellow2014generative}
Ian Goodfellow, Jean Pouget-Abadie, Mehdi Mirza, Bing Xu, David Warde-Farley,
  Sherjil Ozair, Aaron Courville, and Yoshua Bengio.
\newblock Generative adversarial nets.
\newblock In {\em Advances in Neural Information Processing Systems}, pages
  2672--2680, 2014.

\bibitem{Graves2012}
Alex Graves.
\newblock {\em Long Short-Term Memory}.
\newblock Springer Berlin Heidelberg, 2012.

\bibitem{graves2013generating}
Alex Graves.
\newblock Generating sequences with recurrent neural networks.
\newblock {\em Arxiv}, 2013.

\bibitem{ICLRSKETCH}
David Ha and Douglas Eck.
\newblock A neural representation of sketch drawings.
\newblock In {\em International Conference on Learning Representations}, 2018.

\bibitem{hadsell2006dimensionality}
Raia Hadsell, Sumit Chopra, and Yann LeCun.
\newblock Dimensionality reduction by learning an invariant mapping.
\newblock In {\em Computer Vision and Pattern Recognition}, pages 1735--1742,
  2006.

\bibitem{han2021dual}
Junlin Han, Mehrdad Shoeiby, Lars Petersson, and Mohammad~Ali Armin.
\newblock Dual contrastive learning for unsupervised image-to-image
  translation.
\newblock In {\em Computer Vision and Pattern Recognition}, pages 746--755,
  2021.

\bibitem{he2016deep}
Kaiming He, Xiangyu Zhang, Shaoqing Ren, and Jian Sun.
\newblock Deep residual learning for image recognition.
\newblock In {\em Proceedings of the IEEE conference on computer vision and
  pattern recognition}, pages 770--778, 2016.

\bibitem{huang2022AGTGAN}
Hongxiang Huang, Daihui Yang, Gang Dai, Zhen Han, Yuyi Wang, Kin-Man Lam, Fan
  Yang, Shuangping Huang, Yongge Liu, and Mengchao He.
\newblock Agtgan: Unpaired image translation for photographic ancient character
  generation.
\newblock In {\em ACM International Conference on Multimedia}, pages
  5456--5467, 2022.

\bibitem{RD-GAN}
Yaoxiong Huang, Mengchao He, Lianwen Jin, and Yongpan Wang.
\newblock Rd-gan: few/zero-shot chinese character style transfer via radical
  decomposition and rendering.
\newblock In {\em European Conference on Computer Vision}, pages 156--172,
  2020.

\bibitem{jiang2019scfont}
Yue Jiang, Zhouhui Lian, Yingmin Tang, and Jianguo Xiao.
\newblock Scfont: Structure-guided chinese font generation via deep stacked
  networks.
\newblock In {\em AAAI conference on Artificial Intelligence}, pages
  4015--4022, 2019.

\bibitem{kang2020ganwriting}
Lei Kang, Pau Riba, Yaxing Wang, Mar{\c{c}}al Rusinol, Alicia Forn{\'e}s, and
  Mauricio Villegas.
\newblock Ganwriting: content-conditioned generation of styled handwritten word
  images.
\newblock In {\em European Conference on Computer Vision}, pages 273--289,
  2020.

\bibitem{kingma2015adam}
Diederik~P Kingma and Jimmy Ba.
\newblock Adam: A method for stochastic optimization.
\newblock In {\em International Conference on Learning Representations}, 2015.

\bibitem{kong2017handwritten}
Weirui Kong and Bicheng Xu.
\newblock Handwritten chinese character generation via conditional neural
  generative models.
\newblock In {\em Advances in Neural Information Processing Systems}, pages
  4--7, 2017.

\bibitem{kong2022look}
Yuxin Kong, Canjie Luo, Weihong Ma, Qiyuan Zhu, Shenggao Zhu, Nicholas Yuan,
  and Lianwen Jin.
\newblock Look closer to supervise better: One-shot font generation via
  component-based discriminator.
\newblock In {\em Computer Vision and Pattern Recognition}, pages 13482--13491,
  2022.

\bibitem{kotani2020generating}
Atsunobu Kotani, Stefanie Tellex, and James Tompkin.
\newblock Generating handwriting via decoupled style descriptors.
\newblock In {\em European Conference on Computer Vision}, pages 764--780,
  2020.

\bibitem{lian2016automatic}
Zhouhui Lian, Bo Zhao, and Jianguo Xiao.
\newblock Automatic generation of large-scale handwriting fonts via style
  learning.
\newblock In {\em SIGGRAPH Asia Technical Briefs}, pages 1--4, 2016.

\bibitem{lin2015complete}
Jeng-Wei Lin, Chian-Ya Hong, Ray-I Chang, Yu-Chun Wang, Shu-Yu Lin, and
  Jan-Ming Ho.
\newblock Complete font generation of chinese characters in personal
  handwriting style.
\newblock In {\em International Performance Computing and Communications
  Conference}, pages 1--5, 2015.

\bibitem{lin2007style}
Zhouchen Lin and Liang Wan.
\newblock Style-preserving english handwriting synthesis.
\newblock {\em Pattern Recognition}, 40(7):2097--2109, 2007.

\bibitem{liu2011casia}
Cheng-Lin Liu, Fei Yin, Da-Han Wang, and Qiu-Feng Wang.
\newblock Casia online and offline chinese handwriting databases.
\newblock In {\em International Conference on Document Analysis and
  Recognition}, pages 37--41, 2011.

\bibitem{liu2022xmp}
Wei Liu, Fangyue Liu, Fei Ding, Qian He, and Zili Yi.
\newblock Xmp-font: Self-supervised cross-modality pre-training for few-shot
  font generation.
\newblock In {\em Computer Vision and Pattern Recognition}, pages 7905--7914,
  2022.

\bibitem{luo2022slogan}
Canjie Luo, Yuanzhi Zhu, Lianwen Jin, Zhe Li, and Dezhi Peng.
\newblock Slogan: Handwriting style synthesis for arbitrary-length and
  out-of-vocabulary text.
\newblock {\em IEEE Transactions on Neural Networks and Learning Systems},
  2022.

\bibitem{matsumoto2001collection}
Kaoru Matsumoto, Takahiro Fukushima, and Masaki Nakagawa.
\newblock Collection and analysis of on-line handwritten japanese character
  patterns.
\newblock In {\em International Conference on Document Analysis and
  Recognition}, pages 496--500, 2001.

\bibitem{pan2022hilo}
Zizheng Pan, Jianfei Cai, and Bohan Zhuang.
\newblock Fast vision transformers with hilo attention.
\newblock In {\em NeurIPS}, 2022.

\bibitem{park2021few}
Song Park, Sanghyuk Chun, Junbum Cha, Bado Lee, and Hyunjung Shim.
\newblock Few-shot font generation with localized style representations and
  factorization.
\newblock In {\em AAAI Conference on Artificial Intelligence}, pages
  2393--2402, 2021.

\bibitem{park2020contrastive}
Taesung Park, Alexei~A Efros, Richard Zhang, and Jun-Yan Zhu.
\newblock Contrastive learning for unpaired image-to-image translation.
\newblock In {\em European Conference on Computer Vision}, pages 319--345,
  2020.

\bibitem{qiu2021source}
Zhen Qiu, Yifan Zhang, Hongbin Lin, Shuaicheng Niu, Yanxia Liu, Qing Du, and
  Mingkui Tan.
\newblock Source-free domain adaptation via avatar prototype generation and
  adaptation.
\newblock In {\em International Joint Conference on Artificial Intelligence},
  2021.

\bibitem{ren2022learning}
Xuanchi Ren, Tao Yang, Yuwang Wang, and Wenjun Zeng.
\newblock Learning disentangled representation by exploiting pretrained
  generative models: A contrastive learning view.
\newblock In {\em International Conference on Learning Representations}, 2022.

\bibitem{sketchformer}
Leo Sampaio~Ferraz Ribeiro, Tu Bui, John Collomosse, and Moacir Ponti.
\newblock Sketchformer: Transformer-based representation for sketched
  structure.
\newblock In {\em Computer Vision and Pattern Recognition}, pages 14153--14162,
  2020.

\bibitem{tang2021write}
Shusen Tang and Zhouhui Lian.
\newblock Write like you: Synthesizing your cursive online chinese handwriting
  via metric-based meta learning.
\newblock {\em Computer Graphics Forum}, 40(2):141--151, 2021.

\bibitem{tang2019fontrnn}
Shusen Tang, Zeqing Xia, Zhouhui Lian, Yingmin Tang, and Jianguo Xiao.
\newblock Fontrnn: Generating large-scale chinese fonts via recurrent neural
  network.
\newblock {\em Computer Graphics Forum}, 38(7):567--577, 2019.

\bibitem{zi2zi}
Yuchen Tian.
\newblock zi2zi:master chinese calligraphy with conditional adversarial
  networks.
\newblock \url{https://github.com/kaonashi-tyc/zi2zi}, 2017.

\bibitem{TianKI20}
Yonglong Tian, Dilip Krishnan, and Phillip Isola.
\newblock Contrastive representation distillation.
\newblock In {\em International Conference on Learning Representations}, 2020.

\bibitem{deepwritesyn}
Ruben Tolosana, Paula Delgado-Santos, Andres Perez-Uribe, Ruben Vera-Rodriguez,
  Julian Fierrez, and Aythami Morales.
\newblock Deepwritesyn: On-line handwriting synthesis via deep short-term
  representations.
\newblock In {\em AAAI Conference on Artificial Intelligence}, pages 600--608,
  2021.

\bibitem{vaswani2017attention}
Ashish Vaswani, Noam Shazeer, Niki Parmar, Jakob Uszkoreit, Llion Jones,
  Aidan~N Gomez, {\L}ukasz Kaiser, and Illia Polosukhin.
\newblock Attention is all you need.
\newblock In {\em Advances in Neural Information Processing Systems}, pages
  5998--6008, 2017.

\bibitem{wang2018cosface}
Hao Wang, Yitong Wang, Zheng Zhou, Xing Ji, Dihong Gong, Jingchao Zhou, Zhifeng
  Li, and Wei Liu.
\newblock Cosface: Large margin cosine loss for deep face recognition.
\newblock In {\em Computer Vision and Pattern Recognition}, pages 5265--5274,
  2018.

\bibitem{wang2002learning}
Jue Wang, Chenyu Wu, Ying-Qing Xu, Heung-Yeung Shum, and Liang Ji.
\newblock Learning-based cursive handwriting synthesis.
\newblock In {\em International Workshop on Frontiers in Handwriting
  Recognition}, pages 157--162, 2002.

\bibitem{williams1989learning}
Ronald~J Williams and David Zipser.
\newblock A learning algorithm for continually running fully recurrent neural
  networks.
\newblock {\em Neural computation}, 1(2):270--280, 1989.

\bibitem{xie2021dg}
Yangchen Xie, Xinyuan Chen, Li Sun, and Yue Lu.
\newblock Dg-font: Deformable generative networks for unsupervised font
  generation.
\newblock In {\em Computer Vision and Pattern Recognition}, pages 5130--5140,
  2021.

\bibitem{yang2019tet}
Shuai Yang, Jiaying Liu, Wenjing Wang, and Zongming Guo.
\newblock Tet-gan: Text effects transfer via stylization and destylization.
\newblock In {\em AAAI Conference on Artificial Intelligence}, pages
  1238--1245, 2019.

\bibitem{yin2013icdar}
Fei Yin, Qiu-Feng Wang, Xu-Yao Zhang, and Cheng-Lin Liu.
\newblock Icdar 2013 chinese handwriting recognition competition.
\newblock In {\em International Conference on Document Analysis and
  Recognition}, pages 1464--1470, 2013.

\bibitem{yin2016synthesizing}
Hang Yin, Patr{\'\i}cia Alves-Oliveira, Francisco~S Melo, Aude Billard, and Ana
  Paiva.
\newblock Synthesizing robotic handwriting motion by learning from human
  demonstrations.
\newblock In {\em International Joint Conference on Artificial Intelligence},
  pages 3530--3537, 2016.

\bibitem{zhang2017drawing}
Xu-Yao Zhang, Fei Yin, Yan-Ming Zhang, Cheng-Lin Liu, and Yoshua Bengio.
\newblock Drawing and recognizing chinese characters with recurrent neural
  network.
\newblock {\em IEEE Transactions on Pattern Analysis and Machine Intelligence},
  40(4):849--862, 2017.

\bibitem{zhang2021unleashing}
Yifan Zhang, Bryan Hooi, Dapeng Hu, Jian Liang, and Jiashi Feng.
\newblock Unleashing the power of contrastive self-supervised visual models via
  contrast-regularized fine-tuning.
\newblock {\em Advances in Neural Information Processing Systems},
  34:29848--29860, 2021.

\bibitem{zhangself}
Yifan Zhang, Bryan Hooi, HONG Lanqing, and Jiashi Feng.
\newblock Self-supervised aggregation of diverse experts for test-agnostic
  long-tailed recognition.
\newblock In {\em Advances in Neural Information Processing Systems}, 2022.

\bibitem{zhang2021deep}
Yifan Zhang, Bingyi Kang, Bryan Hooi, Shuicheng Yan, and Jiashi Feng.
\newblock Deep long-tailed learning: A survey.
\newblock {\em arXiv preprint arXiv:2110.04596}, 2021.

\bibitem{emd}
Yexun Zhang, Ya Zhang, and Wenbin Cai.
\newblock Separating style and content for generalized style transfer.
\newblock In {\em Computer Vision and Pattern Recognition}, pages 8447--8455,
  2018.

\bibitem{zhao2020deep}
Bocheng Zhao, Jianhua Tao, Minghao Yang, Zhengkun Tian, Cunhang Fan, and Ye
  Bai.
\newblock Deep imitator: Handwriting calligraphy imitation via deep attention
  networks.
\newblock {\em Pattern Recognition}, 104:107080, 2020.

\bibitem{zhuang2021new}
Zhenzhou Zhuang, Zonghao Liu, Kin-Man Lam, Shuangping Huang, and Gang Dai.
\newblock A new semi-automatic annotation model via semantic boundary
  estimation for scene text detection.
\newblock In {\em International Conference on Document Analysis and
  Recognition}, pages 257--273, 2021.

\bibitem{zong2014strokebank}
Alfred Zong and Yuke Zhu.
\newblock Strokebank: Automating personalized chinese handwriting generation.
\newblock In {\em AAAI Conference on Artificial Intelligence}, 2014.

\end{thebibliography}
}

\clearpage

\twocolumn[
\begin{@twocolumnfalse}
\begin{spacing}{1.4}
{\centering
\textbf{\Large{Supplementary Materials: \\ Disentangling Writer and Character Styles for Handwriting Generation}}
\par}
\end{spacing}
\vspace{22pt}
\end{@twocolumnfalse}
]
\setcounter{section}{0}
\renewcommand\thesection{\Alph{section}}
\section{Overview}
\label{sec:appendix}
We organize our supplementary material as follows.
\begin{itemize}
	\item In ~\cref{a.1}, we provide more implementation details.
	\item In ~\cref{a.2}, we provide more visualization examples for spectrum analysis of two style representations.
	\item In ~\cref{a.3}, we describe additional related works about handwriting generation and review the works in font generation.
	\item In ~\cref{a.4}, we provide qualitative results of offline Chinese handwriting generation with a comparison to previous state-of-the-art works.
	\item In ~\cref{a.5}, we study the effect of the sampling ratio $\alpha$ and compare different combination strategies in decoder based on online Chinese handwriting dataset.
       \item In ~\cref{a.6}, we provide the discussions on the format of style inputs.
       \item In ~\cref{a.7}, we conduct failure case analysis.
       \item In ~\cref{a.8}, we report more evaluation metrics on Japanese dataset.
	\item In ~\cref{a.9}, we conudct more experiments on Indic dataset.
        \item In ~\cref{a.10}, we give detailed data representations of online characters.
    \item In ~\cref{a.11}, we describe more details about the pen moving prediction and pen state classification losses.
	\item In ~\cref{a.12}, we show a large number of generated online samples, covering Chinese, Japanese, Indic and English scripts.
	
\end{itemize}

\subsection{More Experimental Details.}\label{a.1}
\subsubsection{Implementation Details of Metrics}\label{a.1.1}
\textbf{DTW} The lower DTW distance, the better quality of the generated characters. For a more robust evaluation metric, we normalize the DTW distance by the length of real online characters to eliminate the effects of different lengths.

\noindent{\textbf{Content and Style Score}} We use the content recognizer and writer identifier to evaluate the Content and Style Score of generated handwritings, respectively. We give the implementation details of the two recognizers below. For the content recognizer~\cite{tang2021write}, we train it on the training set. The optimizer is Adam with the learning rate of 0.001 and the batch size is set to 256. In total, we train four content recognizers on four training sets, i.e., Chinese, Japanese, Indic and English datasets, respectively. \cref{content_rec} summarizes their recognition results on the corresponding test sets. For the writer identifier, we train it on the handwritings belonging to the test writers. Different from the content recognizer receiving a character once, the writer identifier takes 15 characters written by the same person as one input set~\cite{zhao2020deep}. Similarly, we use the Adam optimizer to train four writer identifiers with the batch size of 128, learning rate of 0.001. We report their recognition accuracy in \cref{writer_iden}.

\noindent{\textbf{User Preference Study}} At each time, given a style reference along with several candidates generated by different methods, participants are required to pick up the most similar candidate with the reference. We finally collect 1500 valid responses contributed by 50 volunteers.

\subsubsection{Implementation Details of Robustness Training}\label{a.1.2}
After removing the redundant points of online characters, we follow ~\cite{zhang2017drawing} to normalize the absolute coordinates of points into a standard interval. As mentioned in Sec.~3.3, we define three states ``pen-down", ``pen-up" and ``pen-end" respectively, which are denoted as $m^1, m^2, m^3$. Specifically, pen-down means that the pen is touching the paper now, and the current and following points will be connected by strokes. Pen-up indicates that the pen has just finished a stroke and is to be lifted up. Pen-end means that the pen has finished writing a completed character. It is obvious that pen-end data points are much less than the other two classes. To solve the biased dataset issue,
we pad each online character $Y\small{=}\left[y_1, ..., y_L \right]$ to a fixed length $N_{max}$, where $N_{max}$ is the length of the longest character in our training dataset and $L$ is the length of $Y$, following~\cite{ICLRSKETCH}. As $L$ is usually shorter than $N_{max}$, we set $y_i$ to be $\left(0, 0, 0, 0, 1\right)$, for $i>L$. During training, we set the temperature $\tau\small{=}0.07$. Following \cite{tang2021write}, we use the Gaussian mixture model (GMM) with $m\small{=}20$ bivariate normal distributions, i.e., the final output $O_t \in \mathbb{R}^{123}$.

\begin{table}
  \begin{minipage}{0.47\linewidth}
	\caption{Quantitative evaluations of four content recognizers on four datasets.}
 \vspace{-0.2in}
	\label{content_rec}
    \begin{center}
    \begin{threeparttable} 
	\begin{tabular}{lc}\toprule
    Datasets & Acc.$(\%)$\\ \midrule 
     Chinese~\cite{yin2013icdar} &  95.43 \\
     Japanese~\cite{matsumoto2001collection} & 93.61 \\
     Indic\textsuperscript{\ref{Indic_dataset}} & 94.48 \\
     English~\cite{yin2013icdar} & 80.12 \\
    \bottomrule
	\end{tabular} 
    \end{threeparttable}
    \end{center}
  \end{minipage}
  \hfill
  \begin{minipage}{0.47\linewidth}
	\caption{Quantitative evaluations of four writer identifiers on four datatsets.}
 \vspace{-0.2in}
	\label{writer_iden}
    \begin{center}
    \begin{threeparttable} 
	\begin{tabular}{lc}\toprule
    Datasets & Acc.$(\%)$\\ \midrule 
     Chinese~\cite{yin2013icdar} &  99.98 \\
     Japanese~\cite{matsumoto2001collection} & 99.64 \\
     Indic\textsuperscript{\ref{Indic_dataset}} & 72.54 \\
     English~\cite{yin2013icdar} & 20.57 \\
    \bottomrule
	\end{tabular} 
    \end{threeparttable}
    \end{center}
  \end{minipage}
  \vspace{-0.2in}
 \end{table}

\subsubsection{Implementation Details of Baseline Methods}\label{a.1.3}
\noindent
\textbf{Drawing\&FontRNN} As mentioned in Sec.~4.1, we re-implement the variants of Drawing~\cite{zhang2017drawing} and FontRNN~\cite{tang2019fontrnn} by adding a style branch proposed in DeepImitator~\cite{zhao2020deep}. Specifically, the variants first leverage the CNN encoder~\cite{zhao2020deep} to extract the style vector from the style images, then, following DeepImitator~\cite{zhao2020deep}, they concatenate the obtained vector with the desired character embedding, which is finally fed into their decoder to generate stylized online handwritings.

\noindent
\textbf{WriteLikeYou-v1} adopts the CNN backbone~\cite{zhao2020deep} as its content and style encoder.

\noindent
\textbf{WriteLikeYou-v2} employs the CNN-Transformer architecture as its content and style encoder. Each encoder is a sequential combination of a standard Resnet18~\cite{he2016deep} and a transformer consisting of 2 standard self-attentions layers~\cite{vaswani2017attention}.

\subsubsection{Implementation Details of DTW Matrix}\label{a.1.4}
As mentioned in Sec.~4.2, we generate two groups of characters $\{{\mathbf a}^i\}_{i\small{=}1}^T$ and $\{{\mathbf b}^j\}_{j\small{=}1}^T$ using different style inputs, where $T$ is the number of test writers, ${\mathbf a}^i\small{=}\left[a_1, ..., a_M\right]$ and ${\mathbf b}^j\small{=}\left[b_1, ..., b_M\right]$ denote the $M$ characters belonging to the writer $w_i$ and $w_j$, respectively. Next, we formulate the average DTW distance between ${\mathbf a}^i$ and ${\mathbf b}^j$ as:
\begin{equation}\label{avg_dtw}
d_{ave}({\mathbf a}^i, {\mathbf b}^j)\small{=}\frac{1}{M}\sum_{m\small{=}1}^{M}d(a_m, b_m),
\end{equation}
where $d(\cdot, \cdot)$ is the DTW distance between two characters. Finally, we denote the DTW matrix $\mathbf C\small{=}(c_{ij}) \in \mathbb{R}^{T \times T} $, where $c_{ij}$ can be formulate as:
\begin{equation}\label{dtw_matrix}
c_{ij}\small{=}d_{ave}({\mathbf a}^i, {\mathbf b}^j).
\end{equation} In particular, when $i\small{=}j$, $c_{ij}$ indicates the average DTW distance between generated characters using different style references belonging to the same person.

\subsubsection{Dataset Details}\label{a.1.5}
\textbf{Japanese Dataset} TUAT HANDS~\cite{matsumoto2001collection} database contains about 3 million online handwritten Japanese characters belonging to 271 writers. We randomly select 216 writers for training and 55 writers for testing. Similarly, we use the Ramer–Douglas–Peucker algorithm ($\epsilon\small{=}2$) to preprocess the online characters. After simplification, the maximum sequence length of characters reaches 770, which is a trouble for training RNN~\cite{bengio1994learning}. For a fair comparison with the previous RNN-based works
~\cite{zhao2020deep}, we drop characters with points more than 150, accounting for about $2\%$ of the total datasets~\cite{tang2021write}. After that, the average length of characters is shortened to 68. We render style images from processed online characters and use easily obtainable printed font as content references.

\noindent{\textbf{Indic Dataset}}
Tamil dataset\footnote{\url{http://lipitk.sourceforge.net/datasets/tamilchardata.htm}\label{Indic_dataset}} consists of samples of 156 Indic character classes written by 169 people, which offers an official train set and test set, i.e., 117 writers for training and 52 writers for testing. Similarly, we remove the redundant points of characters via Ramer–Douglas–Peucker algorithm ($\epsilon\small{=}2$) and discard characters with points more than 150. After that, the average sequence length of characters are reduced to 88. We use online Indic characters to render style images. As for content references, we use character embeddings~\cite{zhang2017drawing} instead of offline images. This is because Tamil encodes characters to special indexes that can not be directly matched with the printed font in UTF-8\footnote{\url{https://www.utf8.com}} Format. Briefly, each character embedding is a latent vector embedded by a class label.

\noindent{\textbf{English Dataset}}
In total, we have 53,248 English characters~\cite{liu2011casia} written by 1,020 persons for training, and 3,120 characters~\cite{yin2013icdar} from 60 writers for testing, where the characters written by each writer cover 52 classes.
Similarly, the Ramer–Douglas–Peucker algorithm ($\epsilon=2$) is adopted to remove redundant points of characters, leading to an average sequence length of 30. We render style images using coordinate points of online
characters and employ printed English font as content images.

\begin{figure*}
    \centering
    \includegraphics[width=0.85\linewidth]{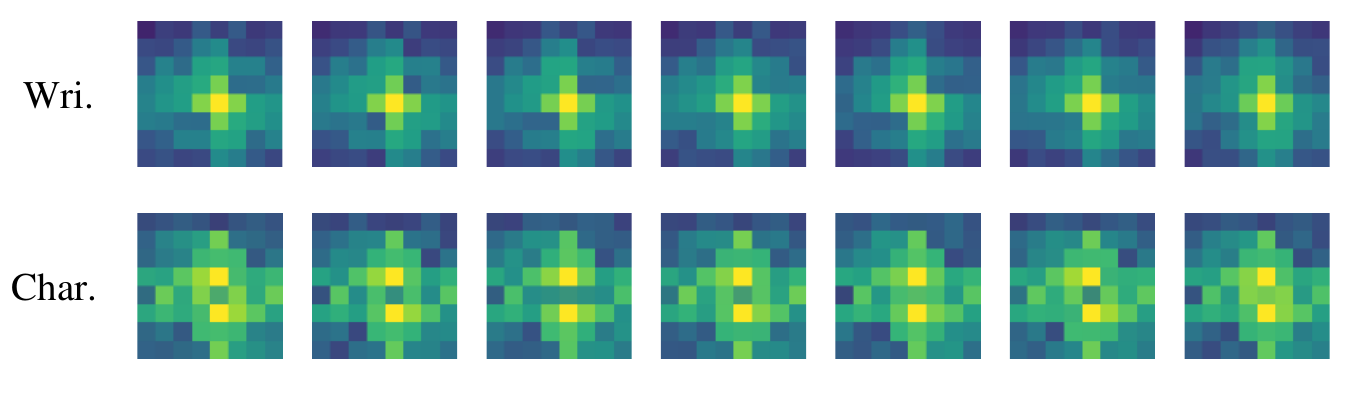}
    \vspace{-0.2in}
    \caption{Frequency magnitude visualizations belong to 7 writers. Each spectrum map is averaged over 100 Chinese character samples.}
    \vspace{-0.1in}
    \label{fft_1}
\end{figure*}

\subsection{More Visualisations on Spectrum Analysis.}\label{a.2}
In \cref{fft_1}, we provide additional frequency magnitude visualizations for writer-wise and character-wise style representations, respectively. Clearly, the results indicate that character-wise styles focus on more high-frequency information, while writer-wise styles mainly pay attention to low-frequency information.

\subsection{More Related Work}\label{a.3}
\textbf{Additional Handwriting Generation Works}
Early traditional approaches are mainly designed to generate Latin characters. Two-step methods~\cite{wang2002learning, lin2007style} generate isolated letters, and then concatenate them to produce a whole word. These methods rely on handcrafted rules and only generate handwritings with limited variations. 

With the rapid development of deep learning, Recurrent Neural Networks (RNNs) and GANs are introduced to generate authentic handwritings~\cite{graves2013generating, fogel2020scrabblegan} conditioned on desired content labels. But these methods are unable to imitate the calligraphic styles of reference samples. DeepWriteSyn~\cite{deepwritesyn}, Sketchformer~\cite{sketchformer} and CoSE~\cite{aksan2020cose} condition the generative process on online handwriting trajectories. Specifically, DeepWriteSYN~\cite{deepwritesyn} introduces the Variational Autoencoder to synthesize realistic forgeries based on given genuine handwritten signatures. Transformer-based methods (\eg, Skecthformer~\cite{sketchformer} and Cose~\cite{aksan2020cose}) adopt the encoder-decoder architecture for reconstructing hand-drawn sketches. However, it is difficult to generalize these methods~\cite{deepwritesyn, sketchformer,aksan2020cose} to multi-style handwriting generation tasks. Specifically, they are confused about using which handwriting style to decorate the given textual content since they lack specific style guidance.

As for handwritten Chinese characters, some previous methods~\cite{zong2014strokebank, lin2015complete,lian2016automatic} extract components (i.e., strokes and radicals) of characters via expert knowledge and then assemble them properly to generate the character. However, these methods rely on hundreds of references, which is labor-intensive.

\textbf{Font Generation}
Generative Adversarial Networks (GANs)~\cite{goodfellow2014generative}
open a new door for font generation and bring amazing performance gains. zi2zi~\cite{zi2zi} regards font generation as an image translation task and achieves diverse font style transfer via a condition GAN. MC-GAN~\cite{MC-GAN} generates the whole set of letters with a consistent style by observing only a few examples via the proposed glyph generation network and texture transfer module. Later, EMD~\cite{emd} and TET-GAN~\cite{yang2019tet} learn the disentangled representations for contents and styles, and thus achieve the unseen style transfer. To further generate high-quality characters, some component-based methods are proposed to take auxiliary annotations (\eg, stroke and radical decomposition) as inputs~\cite{jiang2019scfont, park2021few, liu2022xmp} or supervisions~\cite{RD-GAN, kong2022look}. However, all of the above works do not explicitly consider the geometric deformation of fonts. DG-font~\cite{xie2021dg} introduces a feature deformation skip connection to conduct spatial deformation, thus performing better on cursive characters. Nonetheless, the advanced DG-font struggles to address the large geometric variations, as shown in \cref{all_offline}.

\subsection{Offline Chinese Handwriting Generation.}\label{a.4}
\textbf{Experimental Setting.} To demonstrate the superiority of the proposed offline-to-offline handwriting generation framework, we use the offline character images of ICDAR-2013 competition database~\cite{yin2013icdar}, which contains 60 writers and 3755 different Chinese characters for each writer. We randomly select $80\%$ of the entire dataset as the training set, and the remaining $20\%$ as the test set. As for content images, we use the popular average Chinese font~\cite{jiang2019scfont}. In our experiments, we resize input images to $64\times64$. We insert an extra ornamentation network~\cite{xie2021dg} following the proposed SDT to constitute our offline handwriting generation method. More specifically, our offline method adopts the following pipeline: first generating online handwritings with large shape changes conditioned on input images via the proposed SDT, and rendering offline character images by connecting coordinate points in generated handwriting trajectories, finally decorating the offline characters with realistic stroke width, ink-blot, etc. via the ornamentation network~\cite{xie2021dg}.

We compare our offline generation method with popular font generation and handwriting image generation works. Specifically, (1) font generation methods include zi2zi~\cite{zi2zi} and DG-FONT~\cite{xie2021dg}. (2) handwriting image generation methods, such as GANWriting~\cite{kang2020ganwriting} and HWT~\cite{bhunia2021handwriting} are considered compared methods.

\textbf{Qualitative Comparison.}
\cref{all_offline} shows qualitative comparison between our method with four competitors. To ensure fair comparisons, we randomly select source and target characters with the same textual contents. The rows of ``Source'' present standard characters with different content. Each row of ``Target'' presents characters belonging to the same writer. We can observe that the handwritten characters generated by our method (rows of ``Ours'') yield the most similar styles to target images in terms of geometric shape and ink-blot. Besides, serious artifacts (\eg, blur and collapsed character structure) appear on the handwritings generated by zi2zi (rows of ``Zi2zi'') and HWT (rows of ``HWT''). There are different degrees of stroke missing in the handwritings generated by GANWriting (rows of ``GANW.'') and DG-Font (``rows of DG-F.''). Moreover, except our method, other methods struggle to synthesize the stroke width and ink-blot similar to the target characters.

Further, we provide more qualitative results with a comparison to GANWriting and DG-Font in ~\cref{offline_1}-\cref{offline_4}.

\subsection{More Ablation Studies on Online Chinese Handwriting Dataset.}\label{a.5}
\textbf{Effect of sampling ratio $\alpha$.} We conduct ablation studies to explore the effect of sampling ratio $\alpha$ on the test set. \cref{sampling} summarizes the experimental results in terms of the Style Score. From these results, we observe that a relatively low sampling ratio $\alpha$~(i.e., $25\%$) achieves the best performance. The results indicate that the lower $\alpha$ guides our SDT to focus on the more fine-grained style pattern, which contributes to improving the generation performance of the proposed method in terms of the style imitation.

\textbf{Evaluation of different combinations between the content feature $q$ and previous points $\{y_j\}_{j=1}^{t-1}$.}
To evaluate the effect of different combination strategies between $q$ and $\{y_j\}_{j=1}^{t-1}$, we re-implement a variant of our method by concatenating $q$ with each point $y_j \in\{y_j\}_{j=1}^{t-1}$ and compare it with our method on the test set. As presented in ~\cref{combination}, we find that our combination strategy improves the style consistency without decreasing content correctness of the generated results. This indicates that our method is able to draw global dependencies between $q$ and $\{y_j\}_{j=1}^{t-1}$ unlike previous RNN-based methods~\cite{zhao2020deep} that suffer from the forgetting phenomenon~\cite{Graves2012}, which demonstrates the effectiveness of the proposed method.

\begin{table}
 \caption{Effect of sampling ratio $\alpha$. Our SDT works well with a low sampling ratio~($25\%$).}
\vspace{-0.1in}
    \centering
    \scalebox{0.95}{
    \begin{threeparttable} 
	\begin{tabular}{ccccc}\toprule
        Sampling Ratio & 0.25 & 0.50 & 0.75 & 1.00 \\ \midrule 
        Style Score $\uparrow$ & \textbf{94.50} & 92.07 & 91.91 & 91.54 \\
        \bottomrule
	\end{tabular} 
    \end{threeparttable}}
    \label{sampling}
    
 \end{table}

\begin{table}
 \caption{Evaluation of different combinations between $q$ and $\{y_j\}_{j=1}^{t-1}$.}
    \vspace{-0.1in}
    \centering
    \scalebox{0.9}{
    \begin{threeparttable} 
	\begin{tabular}{cccc}\toprule
        Combination & Style Score $\uparrow$ & Content Score$\uparrow$ & DTW $\downarrow$\\ \midrule 
        Concat & 91.61 & 96.95 & 0.8976 \\
        Ours(SDT) & \textbf{94.50} & \textbf{97.04} & \textbf{0.8789} \\
        \bottomrule
	\end{tabular} 
    \end{threeparttable}}
    \label{combination}
    \vspace{-0.1in}
 \end{table}

\begin{table}
 \caption{Effect of different combinations between $q$, $E$ and $G$.}
 \vspace{-0.1in}
    \centering
    \scalebox{0.9}{
    \begin{threeparttable} 
	\begin{tabular}{cccc}\toprule
        Combination & Style Score $\uparrow$ & Content Score$\uparrow$ & DTW $\downarrow$\\ \midrule 
        Concat & 78.12 & 96.88 & 0.8933 \\
        Ours(SDT) & \textbf{94.50} & \textbf{97.04} & \textbf{0.8789} \\
        \bottomrule
	\end{tabular} 
    \end{threeparttable}}
    
    \label{attention}
 \end{table}
 
\textbf{Evaluation of different combinations between the content feature $q$ and style representations, i.e., $E$ and $G$.}
To demonstrate the effectiveness of our attention-based combination strategy (as mentioned in Sec.~3.3) between $q$, $E$ and $H$, we realize a new version of our SDT by directly concatenating $q$ with $E$ and $H$. Specifically, the new version takes previous points as the query vector, which then attends to combined content and style features. The experimental results are reported in~\cref{attention}. From these results, our method obtains better performance in terms of three quantitative metrics, especially a $16.38\%$ improvement in Style Score, embodying the superiority of our SDT.

\begin{figure}[ht]
    \vspace{-0.1in}
    \centering
    \includegraphics[width=0.98\linewidth]{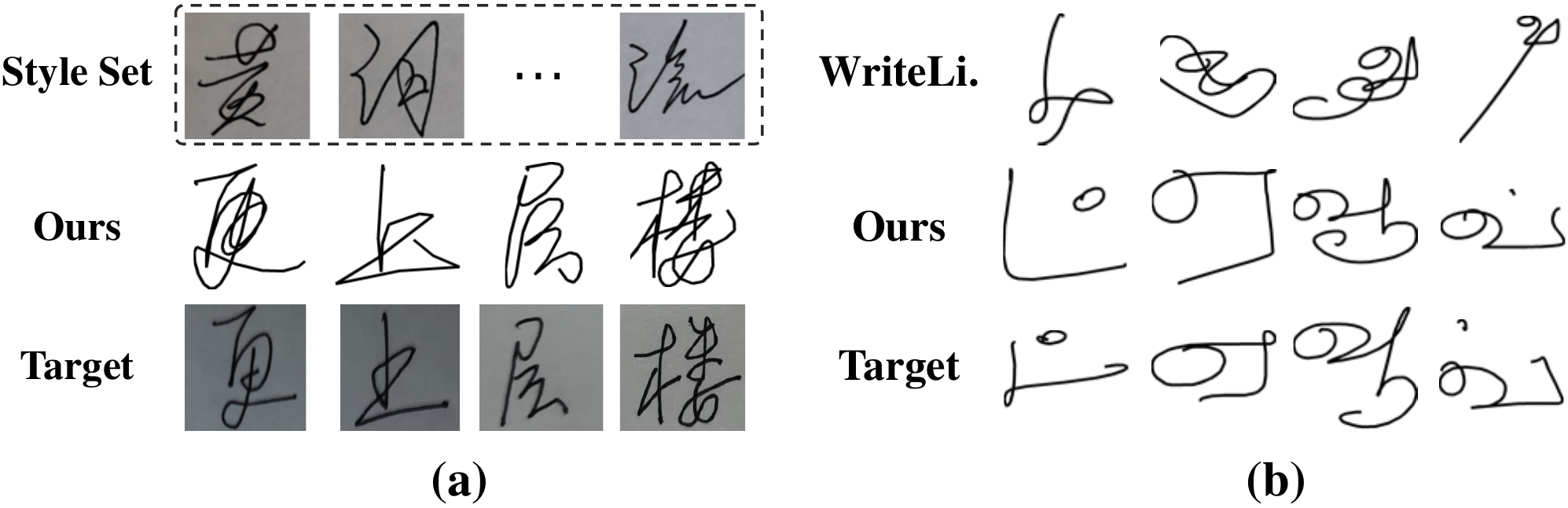}
    \vspace{-0.2in}
    \caption{}
    \label{fig:badcase}
    \vspace{-0.2in}
\end{figure}

\subsection{Discussions on the format of style inputs}\label{a.6}
As described in Sec.1, online trajectories contain more style information (\eg, the order of writing). However, getting trajectories requires users to use specific equipment (\eg, tablets and electric pens), making the methods (that use trajectories as inputs, \eg, WriteLikeYou~\cite{tang2021write}) non-portable in real applications [\textcolor{green}{41}]. Instead, we explore offline images as inputs, which are easier for users to obtain (\eg, through phones). As shown in \cref{fig:badcase} (a), by taking some pictures of the user's handwriting set as the style reference, our method can readily generate the target-stylized online characters. In addition, compared with original WriteLikeYou~\cite{tang2021write}, our method receives less input information, but we still achieve comparable performance with it, in terms of Style Score ($94.50\%$ $vs.$ $92.85\%$) and Content Score ($97.04\%$ $vs.$ $97.92\%$). Overall, our method has better applicability in real scenarios.

\subsection{Analysis of failure cases}\label{a.7}
We provide some failure cases  in \cref{fig:badcase} (b), where our SDT fails to imitate the overall style (e.g., glyph slant and aspect ratios) for some Indic characters. However, thanks to our disentangling scheme, SDT can still capture their detailed style (e.g., stroke location and curvature), while previous methods (e.g.,~WriteLikeYou) cannot imitate the style of target characters at all.

\begin{table}[t]
	\caption{Additional quantitative evaluations of our SDT and competitors on Japanese dataset.}
	\vspace{-0.25in}
	\label{tab:Janpanese}
    \begin{center}
  \scalebox{0.85}{
    \begin{threeparttable} 
	\begin{tabular}{lc}\toprule
    Methods & Style Score $\uparrow$\\ \midrule 
      Drawing~\cite{zhang2017drawing}  & 20.67   \\
      DeepImitator~\cite{zhao2020deep}  & 25.80   \\
       WriteLikeYou-v2~\cite{tang2021write}  & 32.88 \\
       SDT(Ours)  & \textbf{41.85} \\
        \bottomrule
	\end{tabular} 
    \end{threeparttable}}
    \end{center}
    \vspace{-0.3in}
\end{table}

\subsection{More Evaluation metrics on Japanese Dataset.}\label{a.8}
We report the experimental results on Japanese dataset in \cref{tab:Janpanese}, in terms of Style Score.
From these results, our SDT outperforms the second best, i.e., WriteLikeYou-v2~\cite{tang2021write}, by a large margin ($41.85\%$ $vs.$ $32.88\%$), which further demonstrates our method has a better imitation performance in respect of handwriting styles regardless of the script type.

\subsection{More Experiments on Indic Dataset.}\label{a.9}
As described in Sec.~4.3, previous works~\cite{zhang2017drawing,tang2021write} achieve very poor generation results~(\eg, Content Score of 0.02) in Indic scripts, which means generating Indic handwritings with certain contents and specific styles may be too difficult for them. To this end, we reduce the difficulty of the Indic handwriting generation task and condition the generative process only on character contents. Quantitative comparison further demonstrates the superiority of our method. We detail the experimental setting and results below. 

For this new task, we still conduct experiments on Tamil\textsuperscript{\ref{Indic_dataset}} dataset to compare our method with other competitors~(i.e., Drawing~\cite{zhang2017drawing} and WriteLikeYou~\cite{tang2021write}). For a fair comparison, without any style reference, all methods take input as the content reference (i.e., character embeddings\cite{zhang2017drawing}) and aim to synthesize handwritings consistent with the given content. Since Drawing ~\cite{zhang2017drawing} is initially designed for content-conditioned generation, we keep its original architecture. For WriteLikeYou ~\cite{tang2021write} and our SDT, we remove their style and content encoder and directly input character embeddings into their decoder.

We provide the experiment results in \cref{content-condition}. From these results, although Drawing~\cite{zhang2017drawing} and WriteLikeYou~\cite{tang2021write} achieve the better performance than their content-style-conditioned generation settings~(as shown in Sec.~4.3), our SDT still achieves the best results in terms of Content Score and DTW. Besides, compared with the content-conditioned generation, our content-style-conditioned results~(as shown in Sec.~4.3) obtain higher Content Score (i.e, $97.22\%$ $vs.$ $68.50\%$) and better DTW (i.e., $0.7075$ $vs.$ $0.9748$), which demonstrates that the extracted style representations by our method further improve the generation quality in terms of Content Score and DTW.

\begin{table}
 \caption{Quantitative evaluations of our SDT and competitors on content-conditioned generation of Indic handwritings.}
 \vspace{-0.1in}
    \centering
    \scalebox{0.85}{
    \begin{threeparttable} 
	\begin{tabular}{lcc}\toprule
        Methods & Content Score $\uparrow$ & DTW $\downarrow$\\ \midrule 
        Drawing~\cite{zhang2017drawing} & 26.07 & 2.7604 \\
        WriteLikeYou~\cite{tang2021write} & 40.29 & 1.0503 \\
        Ours(SDT)  & \textbf{68.50} & \textbf{0.9748} \\
        \bottomrule
	\end{tabular} 
    \end{threeparttable}}
    \label{content-condition}
    \vspace{-0.15in}
 \end{table}

\subsection{Data Representations of Online Characters}\label{a.10}
Generally, each online character is composed of a sequence of points and can be mathematically represented as $Y\small{=}\left[y_1, ..., y_L \right]$, where $L$ is the length of $Y$. Following~\cite{zhang2017drawing}, each point is a vector with 5 elements $y_t\small{=}\left(\Delta{u_t}, \Delta{v_t}, m_t^1, m_t^2, m_t^3\right)$, where $\left(\Delta{u_t}, \Delta{v_t}\right)$ are the relative offsets from the current point to the previous point and ($m_t^1$-down, $m_t^2$-up, $m_t^3$-end) are three types of pen states, which are mutually exclusive.

\subsection{More Details about the Pen Moving Prediction and Pen State Classification Losses}\label{a.11}
During training, we have the ground-truth point $y_t\small{=}\left(\Delta{u}, \Delta{v}, m_1, m_2, m_3\right)$ and the final output $O_t=(\left\{\hat{\pi}^i, \hat{\mu}_x^i, \hat{\mu}_y^i, \hat{\delta}_x^i, \hat{\delta}_y^i, \hat{\rho}_{x y}^i\right\}_{i=1}^R, \hat{m}_1, \hat{m}_2, \hat{m}_3)$ of our decoder at any time step $t$. Here, ${\hat{\pi}}^i$ is the component weight of different bivariate normal distributions, $\hat{\mu}_x^i$ and $\hat{\mu}_y^i$ are the means of distributions, $\hat{\delta}_x^i$, $\hat{\delta}_y^i$ denotes the standard deviations of distributions and $\hat{\rho}_{x y}^i$ is the covariance, as suggested in~\cite{tang2021write}. Then, the pen moving prediction loss for each time step can be formulated as:
\begin{equation}
   \mathcal{L}_{pre}=\sum_{i=1}^R \hat{\pi}^i \mathcal{N}\left(\Delta u, \Delta v \mid \hat{\mu}_x^i, \hat{\mu}_y^i, \hat{\delta}_x^i, \hat{\delta}_y^i, \hat{\rho}_{x y}^i\right), 
\end{equation}
where $\mathcal{N(\cdot)}$ is the bivariate normal distribution function.

Regarding the pen state classification loss at any time step, we formulate it as follows:
\begin{equation}
\mathcal{L}_{cls}=\sum_{i=1}^3 m_i \log{\hat{m}_i}.
\end{equation}

\subsection{More Online Generation Results.}\label{a.12}
~\cref{online_1}-\cref{other_1} show qualitative comparisons between our proposed SDT and the previous state-of-the-art work WriteLikeYou~\cite{tang2021write} on online multilingual characters generation (\eg, Chinese, Japanese, Indic and English scripts). The results suggest that our method is more competitive in both style imitation and structure preservation of generated multilingual characters.

\clearpage
\begin{figure*}
    \centering
\includegraphics[width=0.75\linewidth]{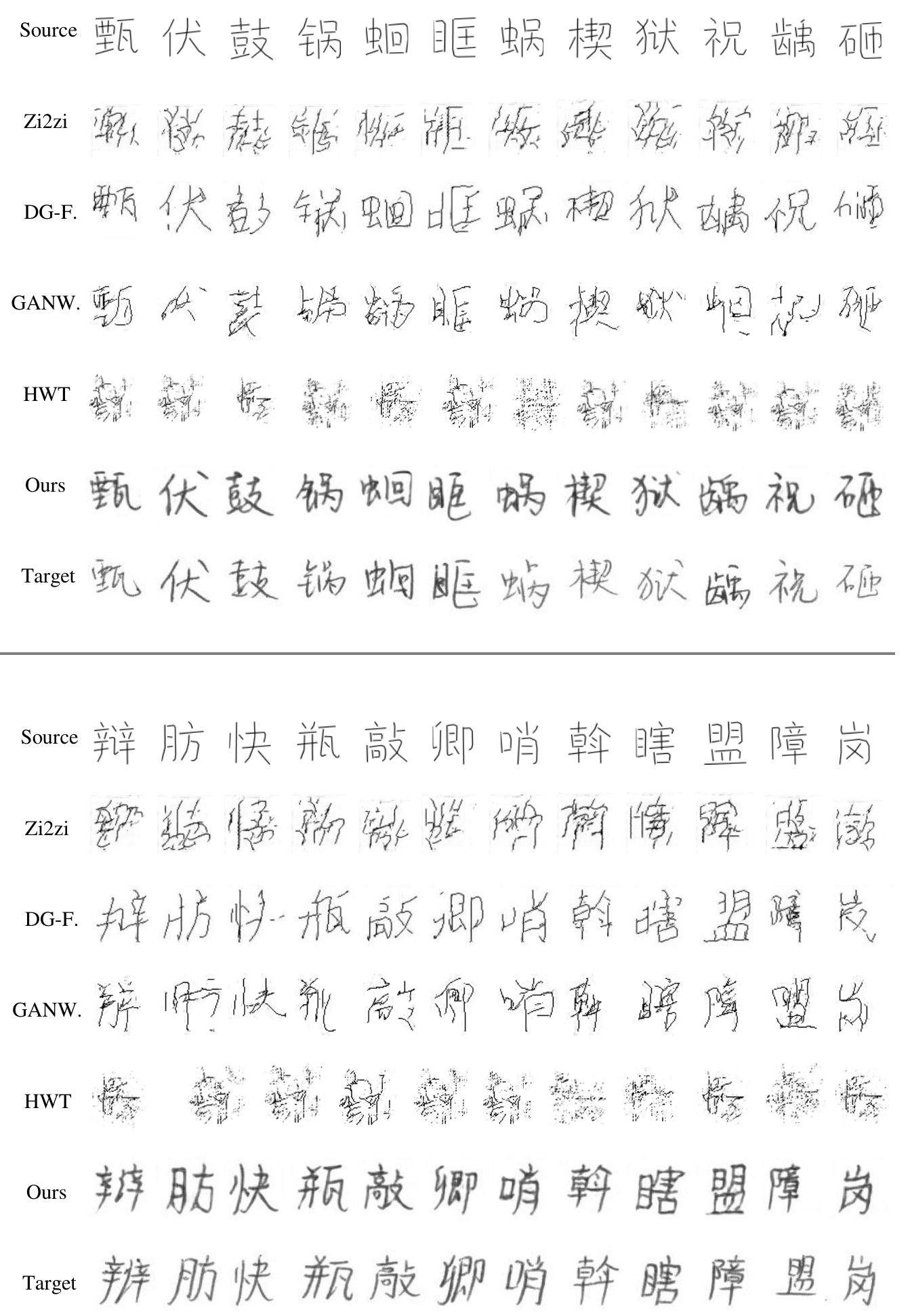}
    \caption{Qualitative comparisons between our proposed SDT with four competitors, including zi2zi~\cite{zi2zi}, DG-FONT~\cite{xie2021dg}, GANWriting~\cite{kang2020ganwriting} and HWT~\cite{bhunia2021handwriting}, on offline handwritten Chinese character generation.}
    \label{all_offline}
\end{figure*}

\begin{figure*}
    \centering
    \includegraphics[width=0.75\linewidth]{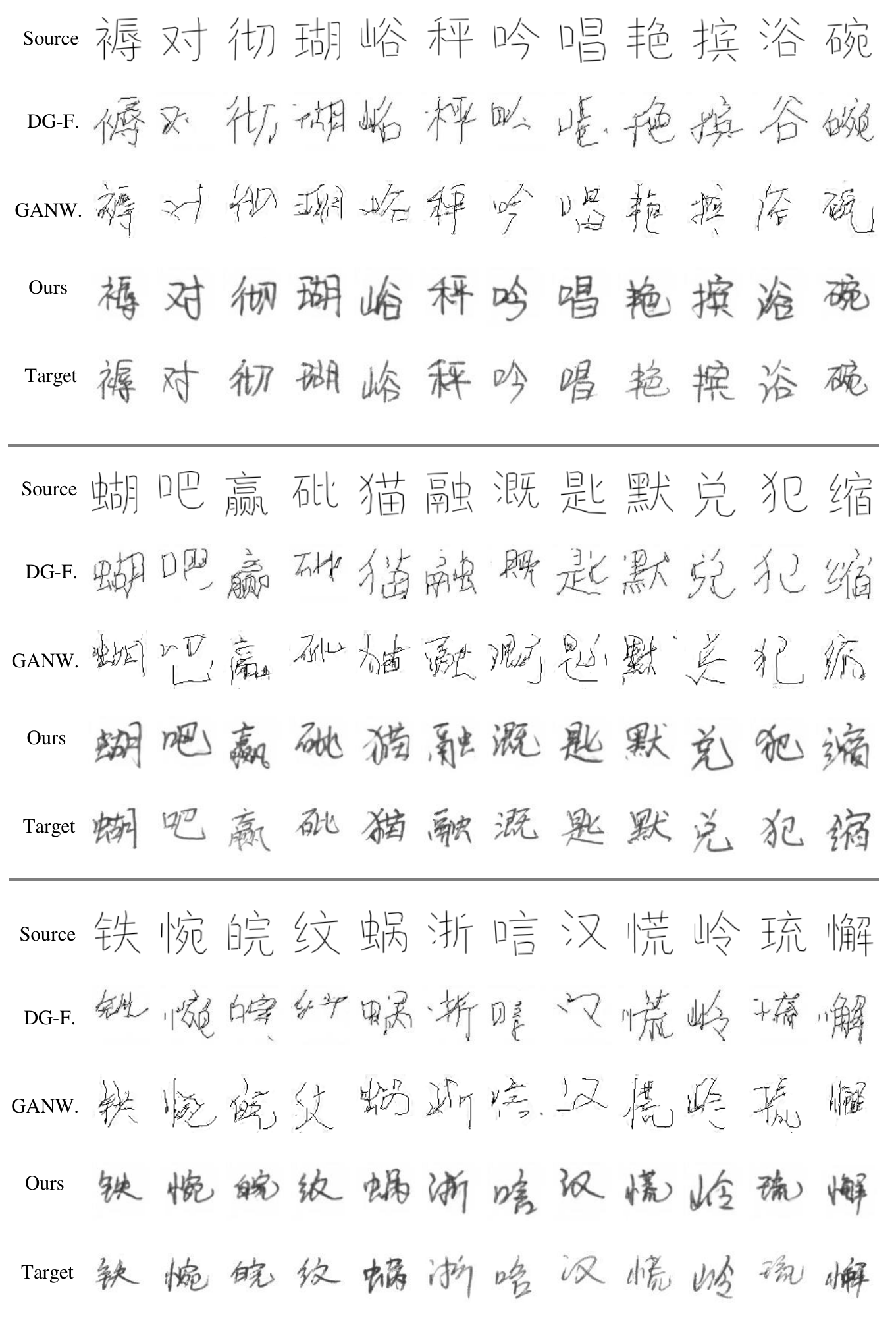}
    \caption{Qualitative comparisons between our proposed SDT with DG-FONT~\cite{xie2021dg} and GANWriting~\cite{kang2020ganwriting}, on offline handwritten Chinese character generation.}
    \label{offline_1}
\end{figure*}

\begin{figure*}
    \centering
    \includegraphics[width=0.75\linewidth]{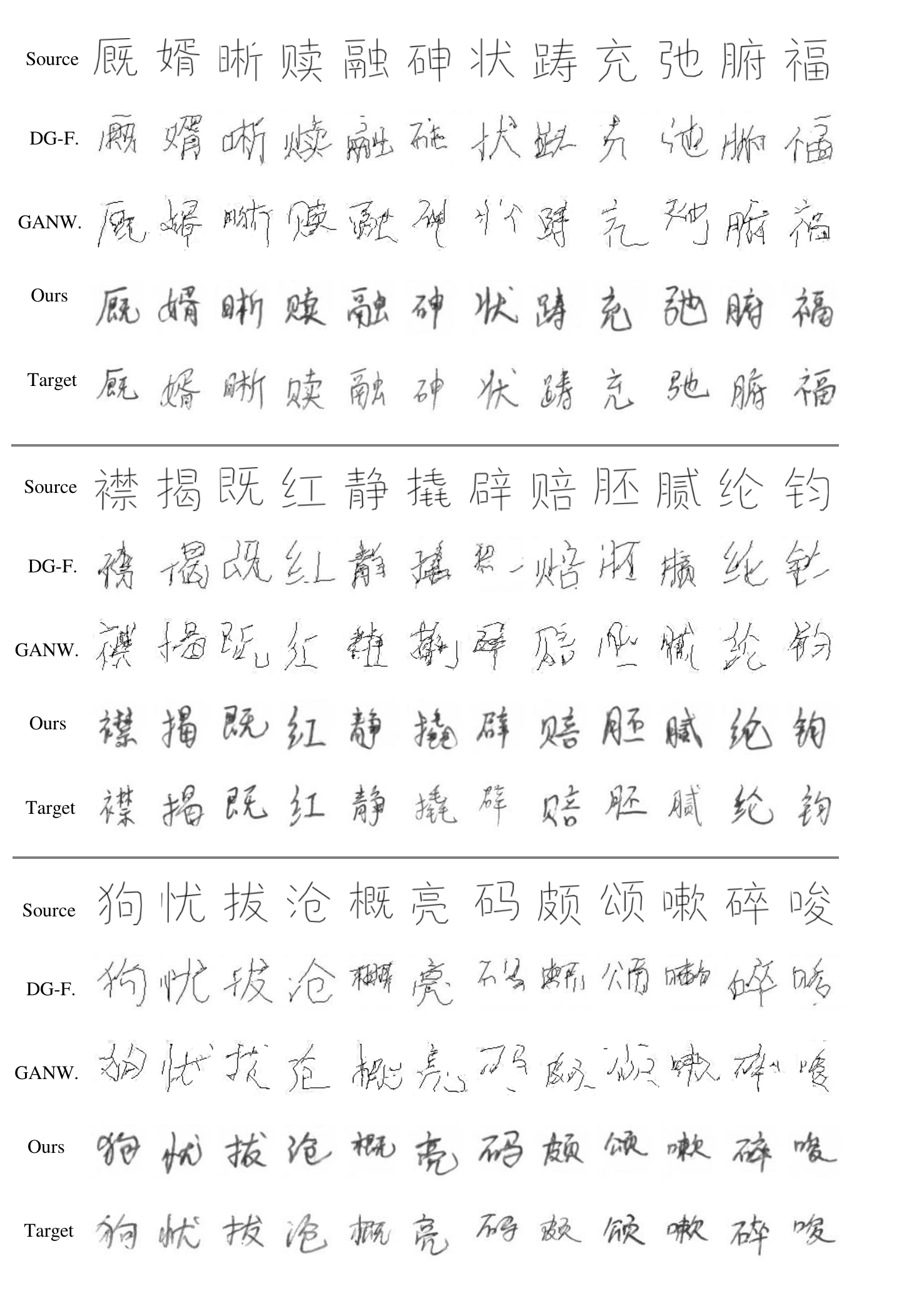}
    \caption{Qualitative comparisons between our proposed SDT with DG-FONT~\cite{xie2021dg} and GANWriting~\cite{kang2020ganwriting}, on offline handwritten Chinese character generation.}
    \label{offline_2}
\end{figure*}

\begin{figure*}
    \centering
    \includegraphics[width=0.75\linewidth]{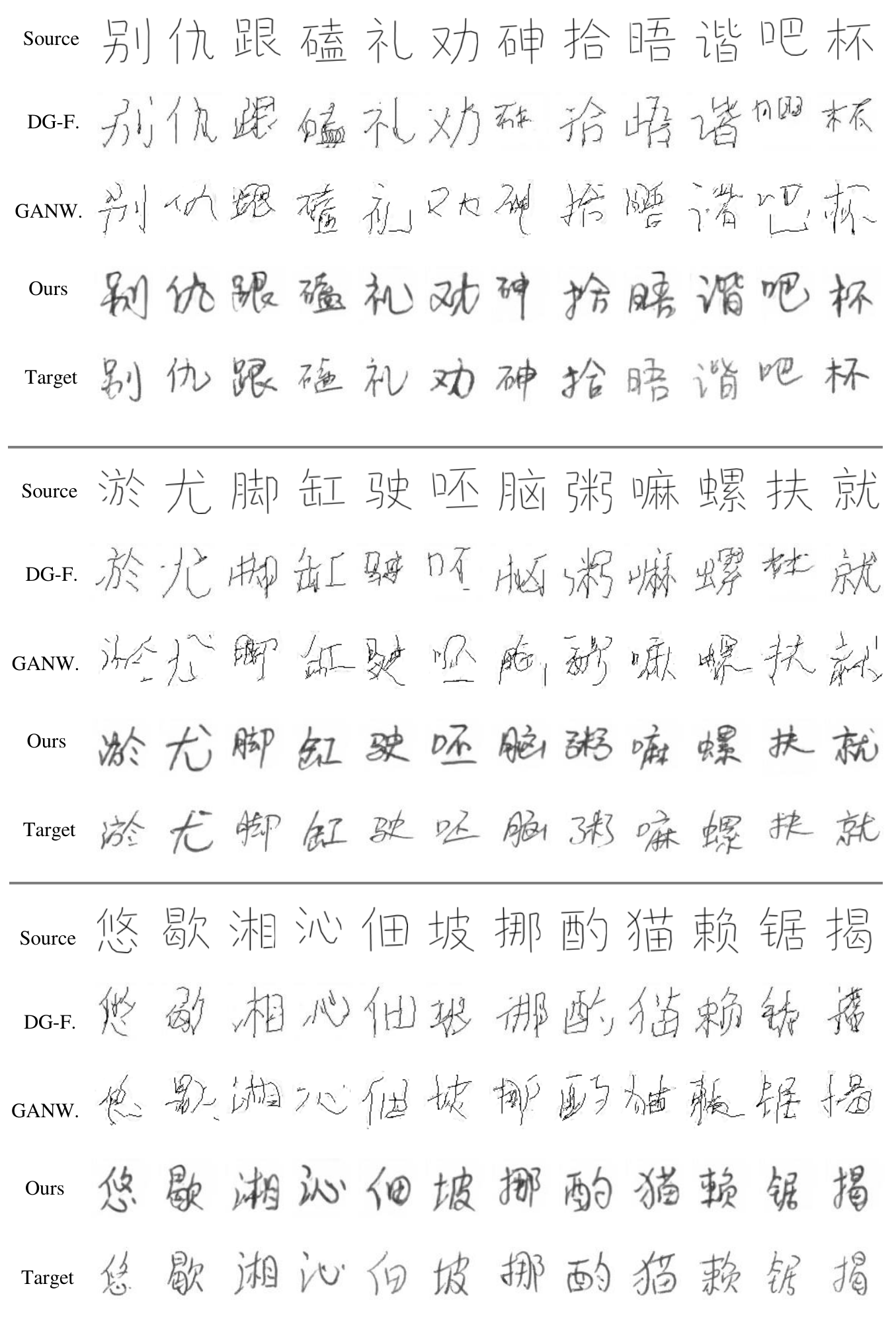}
    \caption{Qualitative comparisons between our proposed SDT with DG-FONT~\cite{xie2021dg} and GANWriting~\cite{kang2020ganwriting}, on offline handwritten Chinese character generation.}
    \label{offline_3}
\end{figure*}

\begin{figure*}
    \centering
    \includegraphics[width=0.75\linewidth]{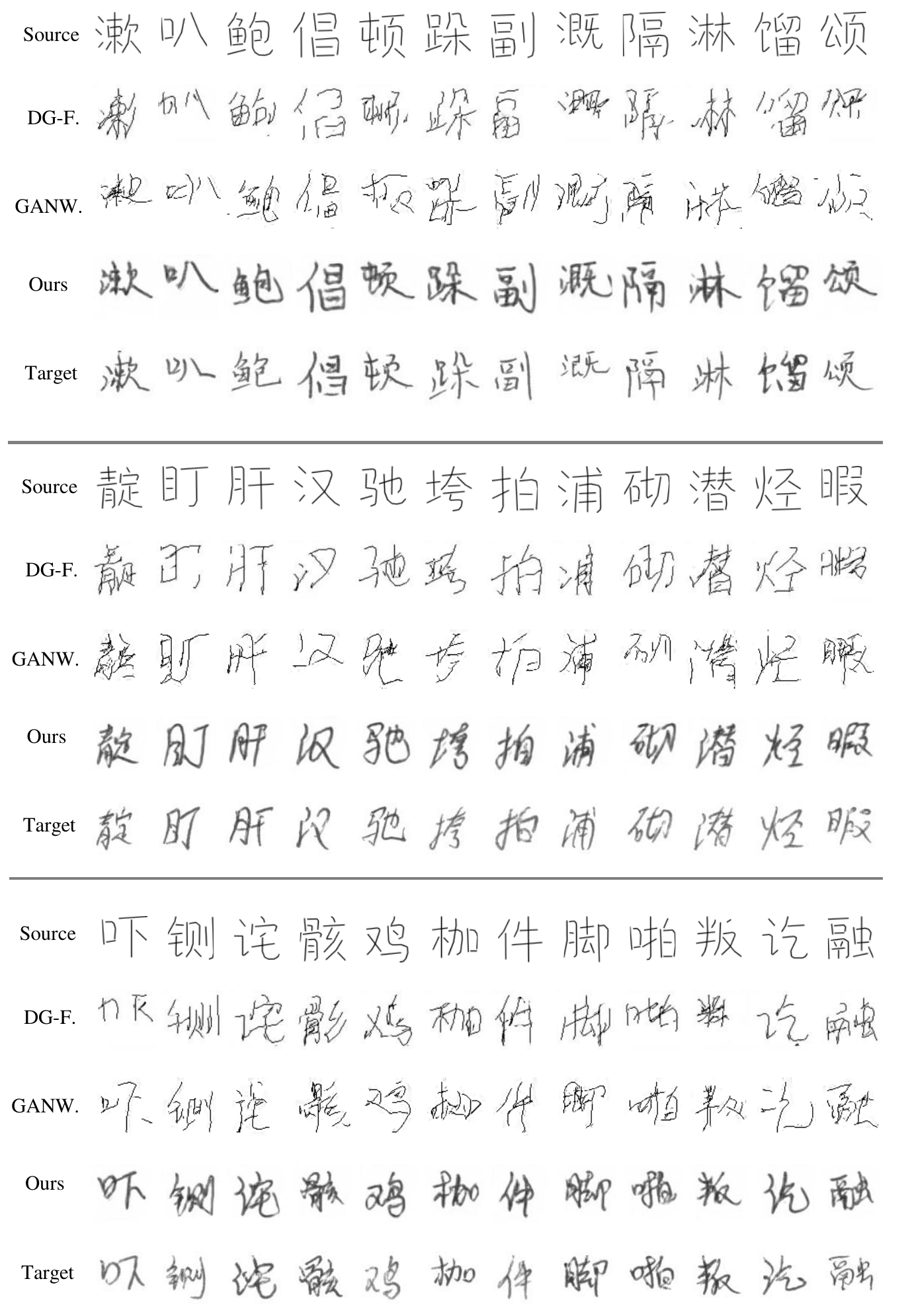}
    \caption{Qualitative comparisons between our proposed SDT with DG-FONT~\cite{xie2021dg} and GANWriting~\cite{kang2020ganwriting}, on offline handwritten Chinese character generation.}
    \label{offline_4}
\end{figure*}

\begin{figure*}
    \centering
    \includegraphics[width=0.75\linewidth]{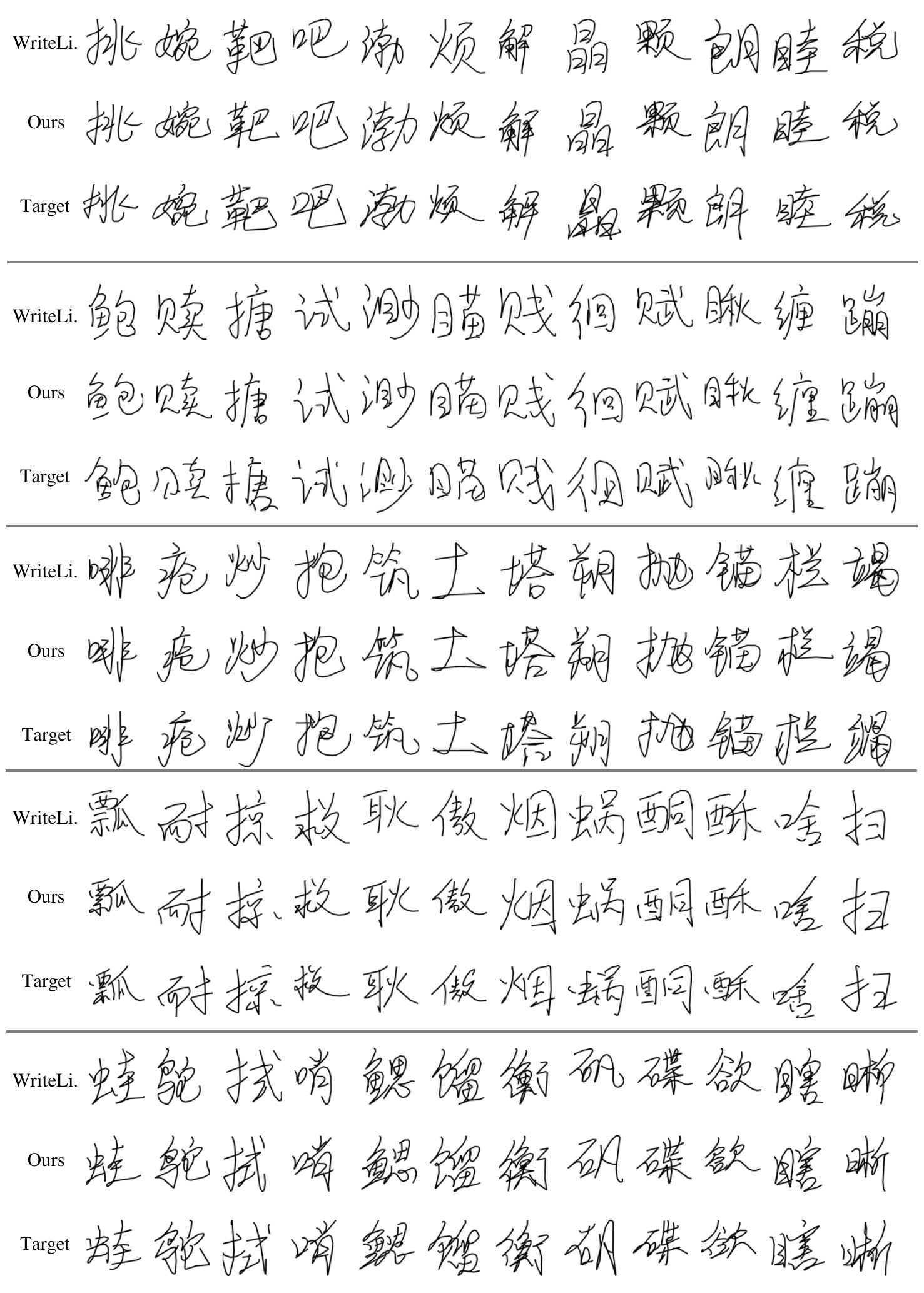}
    \caption{Additional generated online Chinese characters by our method and WriteLikeYou-v2~\cite{tang2021write}.}
    \label{online_1}
    \vspace{+0.8in}
\end{figure*}

\begin{figure*}
    \centering
    \includegraphics[width=0.75\linewidth]{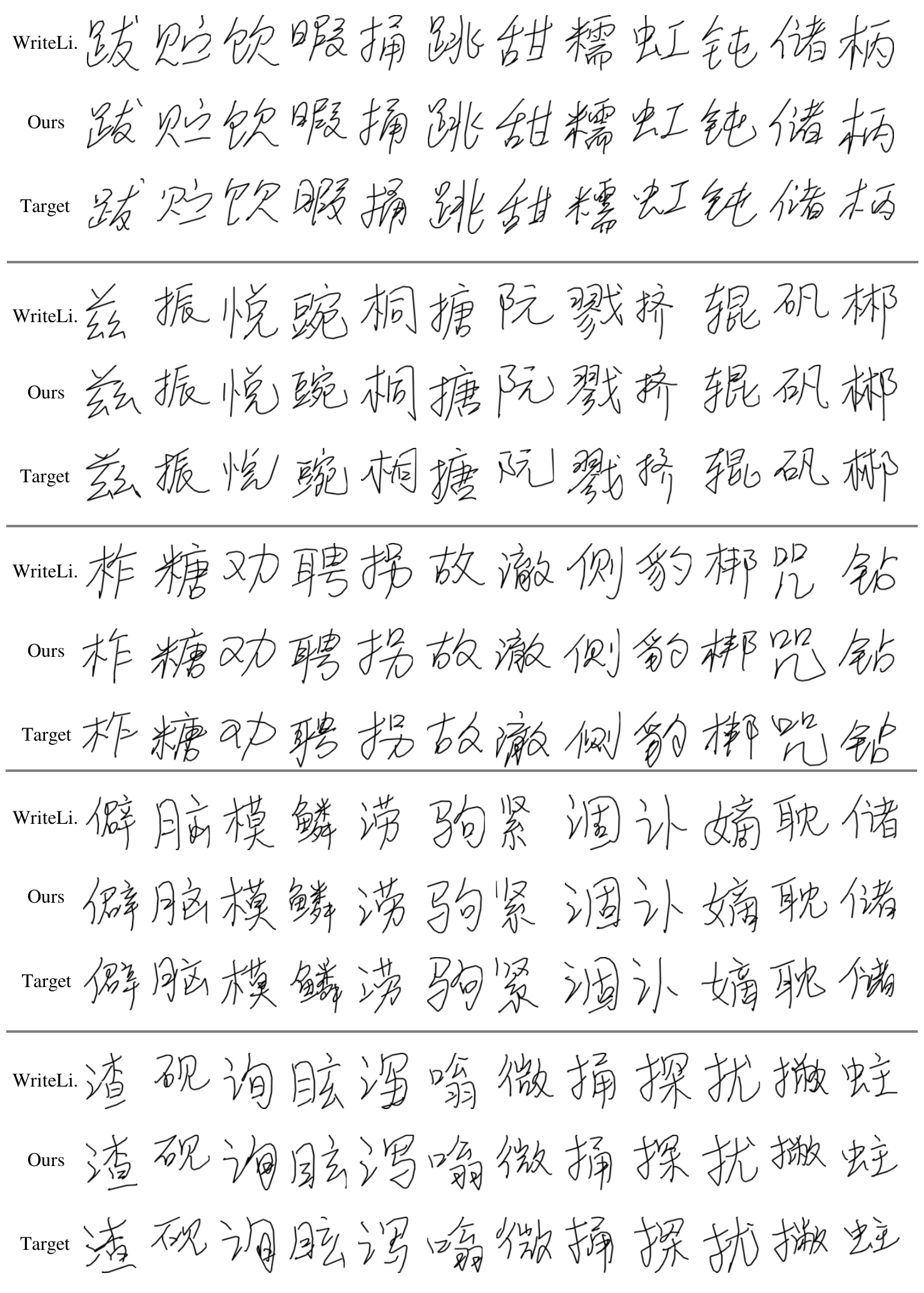}
    \caption{Additional generated online Chinese characters by our method and WriteLikeYou-v2~\cite{tang2021write}.}
    \label{online_2}
    \vspace{+0.8in}
\end{figure*}

\begin{figure*}
    \centering
    \includegraphics[width=0.75\linewidth]{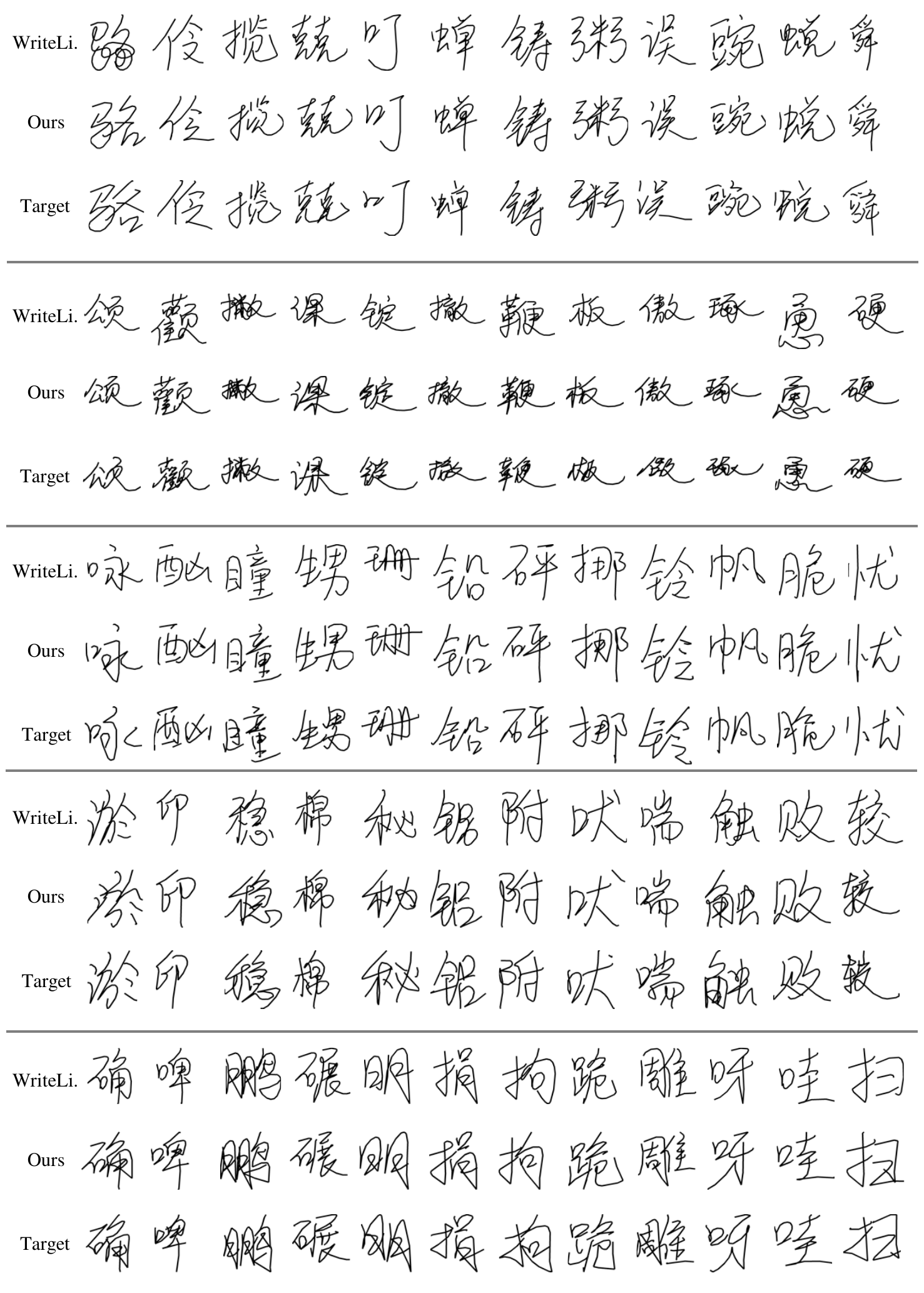}
    \caption{Additional generated online Chinese characters by our method and WriteLikeYou-v2~\cite{tang2021write}.}
    \label{online_3}
    \vspace{+0.8in}
\end{figure*}

\begin{figure*}
    \centering
    \includegraphics[width=0.75\linewidth]{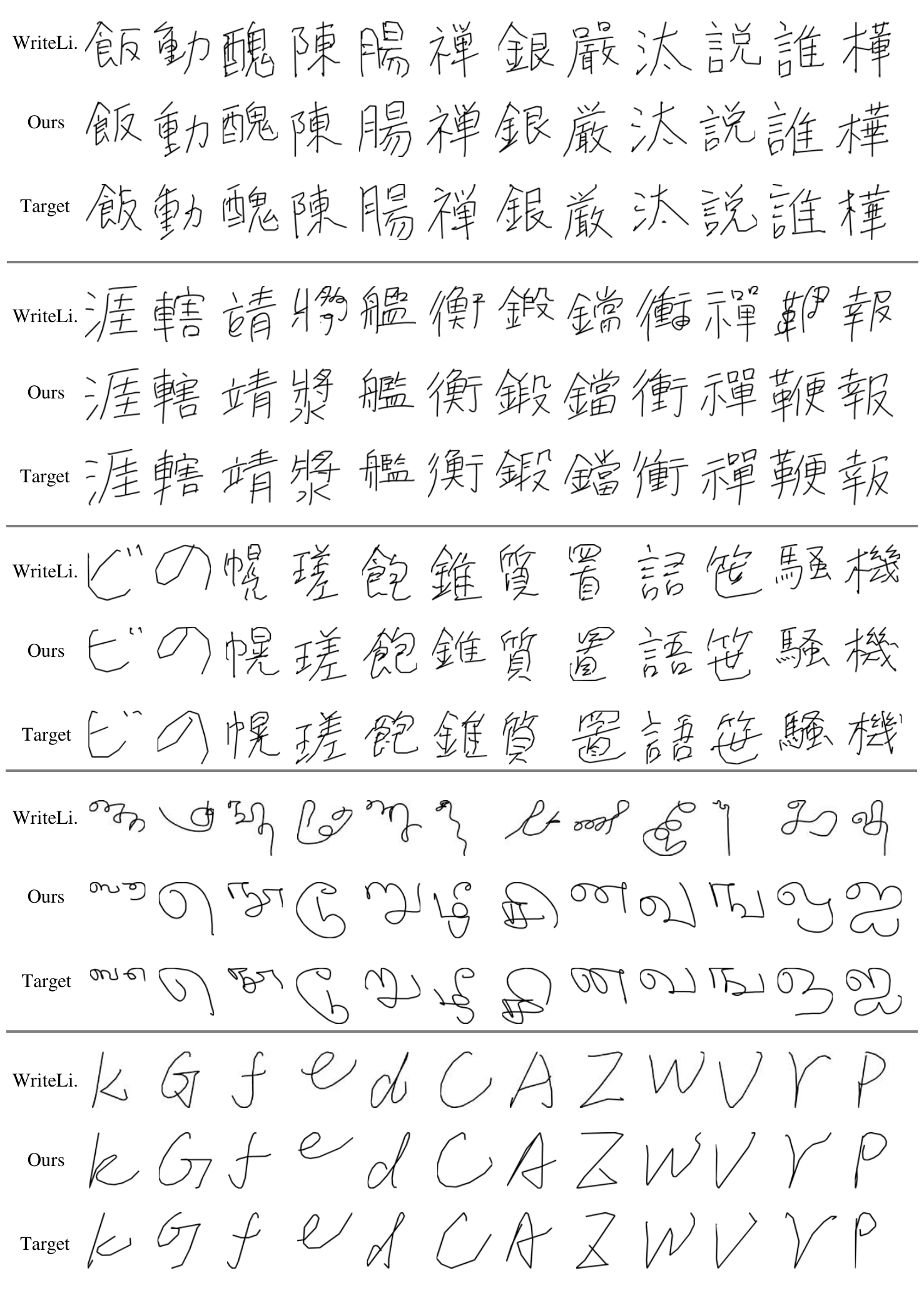}
    \caption{Additional generated online characters, covering Japanese, Indic and English scripts, by our method and WriteLikeYou-v2~\cite{tang2021write}.}
    \label{other_1}
    \vspace{+0.8in}
\end{figure*}

\end{document}